\documentclass[journal]{IEEEtran}
\ifCLASSINFOpdf
\else
\fi
\ifCLASSINFOpdf
   \usepackage[pdftex]{graphicx}
   \graphicspath{{../Figures/}}
   \DeclareGraphicsExtensions{.pdf,.jpeg,.png}
\else
\fi

\usepackage{amsmath}
\usepackage{amssymb}
\usepackage{comment}

\usepackage{amsmath}

\usepackage{array}

\ifCLASSOPTIONcompsoc
 \usepackage[caption=false,font=normalsize,labelfont=sf,textfont=sf]{subfig}
\else
  \usepackage[caption=false,font=footnotesize]{subfig}
\fi
\usepackage{stfloats}
\usepackage{tabularx,booktabs}
\usepackage{multirow}
\usepackage{amssymb}
\usepackage{array}
\usepackage{xcolor}
\usepackage{cite}
\begin{document}
\title{Fine-Grained Sports, Yoga, and Dance Postures Recognition: A Benchmark Analysis}

\author{Asish~Bera,~\IEEEmembership{Member,~IEEE,} ~Mita~Nasipuri,~\IEEEmembership{Life Senior Member,~IEEE}\\ 
Ondrej Krejcar, and
        Debotosh~Bhattacharjee,~\IEEEmembership{Senior Member,~IEEE}
    
\thanks{A. Bera is with the Department of Computer Science and Information Systems, Birla Institute of Technology and Science, Pilani, Pilani Campus,  Rajasthan, 333031, India.  E-mail: (asish.bera@pilani.bits-pilani.ac.in).}
\thanks{M. Nasipuri is with the Department of Computer Science and Engineering, Jadavpur University, Kolkata 32, India.  Email: mitanasipuri@gmail.com}
\thanks{O. Krejcar is with the Center for Basic and Applied Science, Faculty of Informatics and Management, University of Hradec Kralove, Rokitanskeho 62, 500 03 Hradec Kralove, Czech Republic. Email:
ondrej.krejcar@uhk.cz}
\thanks{D. Bhattacharjee is with the Department of Computer Science and Engineering, Jadavpur University, Kolkata 32, India. He is also with the Center for Basic and Applied Science, Faculty of Informatics and Management, University of Hradec Kralove, Rokitanskeho 62, 500 03 Hradec Kralove, Czech Republic.  Email:  debotosh@ieee.org}
}

\markboth{Journal of \LaTeX\ Class Files,~Vol.~00, No.~0, Feb~2023}%
{Bera \MakeLowercase{\textit{et al.}}: Bare Demo of IEEEtran.cls for IEEE Journals}
%

\maketitle

\begin{abstract}
Human body-pose estimation is a complex problem in computer vision.  Recent research interests have been widened specifically on the Sports, Yoga, and Dance (SYD) postures for maintaining health conditions. The SYD pose categories are regarded as a fine-grained image classification task due to the complex movement of body parts. Deep Convolutional Neural Networks (CNNs) have attained significantly improved performance in solving various human body-pose estimation problems. Though decent progress has been achieved in yoga postures recognition using deep learning techniques, fine-grained sports, and dance recognition necessitates ample research attention. However, no benchmark public image dataset with sufficient inter-class and intra-class variations is available yet to address sports and dance postures classification. To solve this limitation, we have proposed two image datasets, one for 102 sport categories and another for 12 dance styles. Two public datasets, Yoga-82 which contains 82 classes and Yoga-107 represents 107 classes are collected for yoga postures. These four SYD datasets are experimented with the proposed deep model, SYD-Net, which integrates a patch-based attention (PbA) mechanism on top of standard backbone CNNs. The PbA module leverages the self-attention mechanism that learns contextual information from a set of uniform and multi-scale patches and emphasizes discriminative features to understand the semantic correlation among patches. Moreover, random erasing data augmentation is applied to improve performance. The proposed SYD-Net has achieved state-of-the-art accuracy on Yoga-82 using five base CNNs. SYD-Net's accuracy on other datasets is remarkable, implying its efficiency. Our Sports-102 and Dance-12 datasets are publicly available at \texttt{https://sites.google.com/view/syd-net/home}. 
\end{abstract}


%
\IEEEpeerreviewmaketitle

\section{Introduction}

\begin{figure}[h]
\centering
\subfloat{ 
\includegraphics[width=0.15\linewidth, height= 1.5 cm]{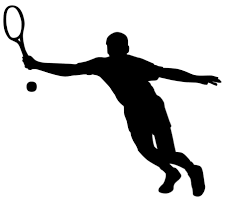} \hfill
\includegraphics[width=0.15\linewidth, height=1.5 cm]{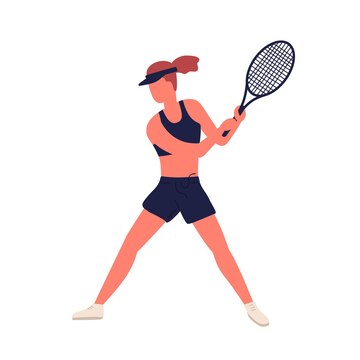} \hfill
\includegraphics[width=0.15\linewidth, height=1.5 cm]{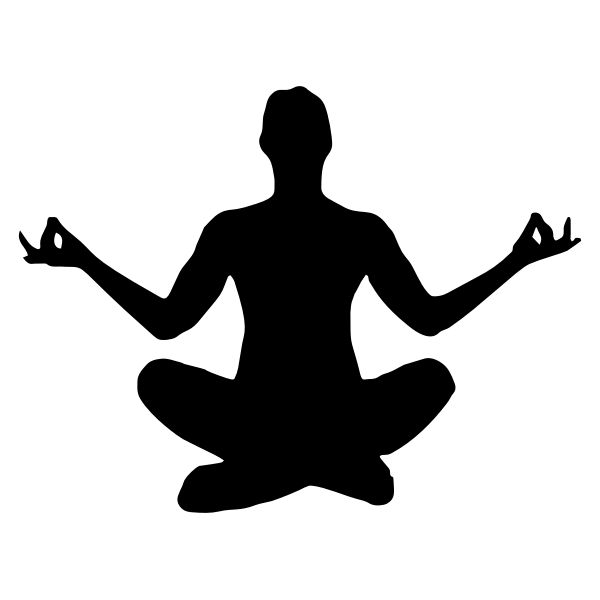} 
\includegraphics[width=0.15\linewidth, height= 1.5 cm]{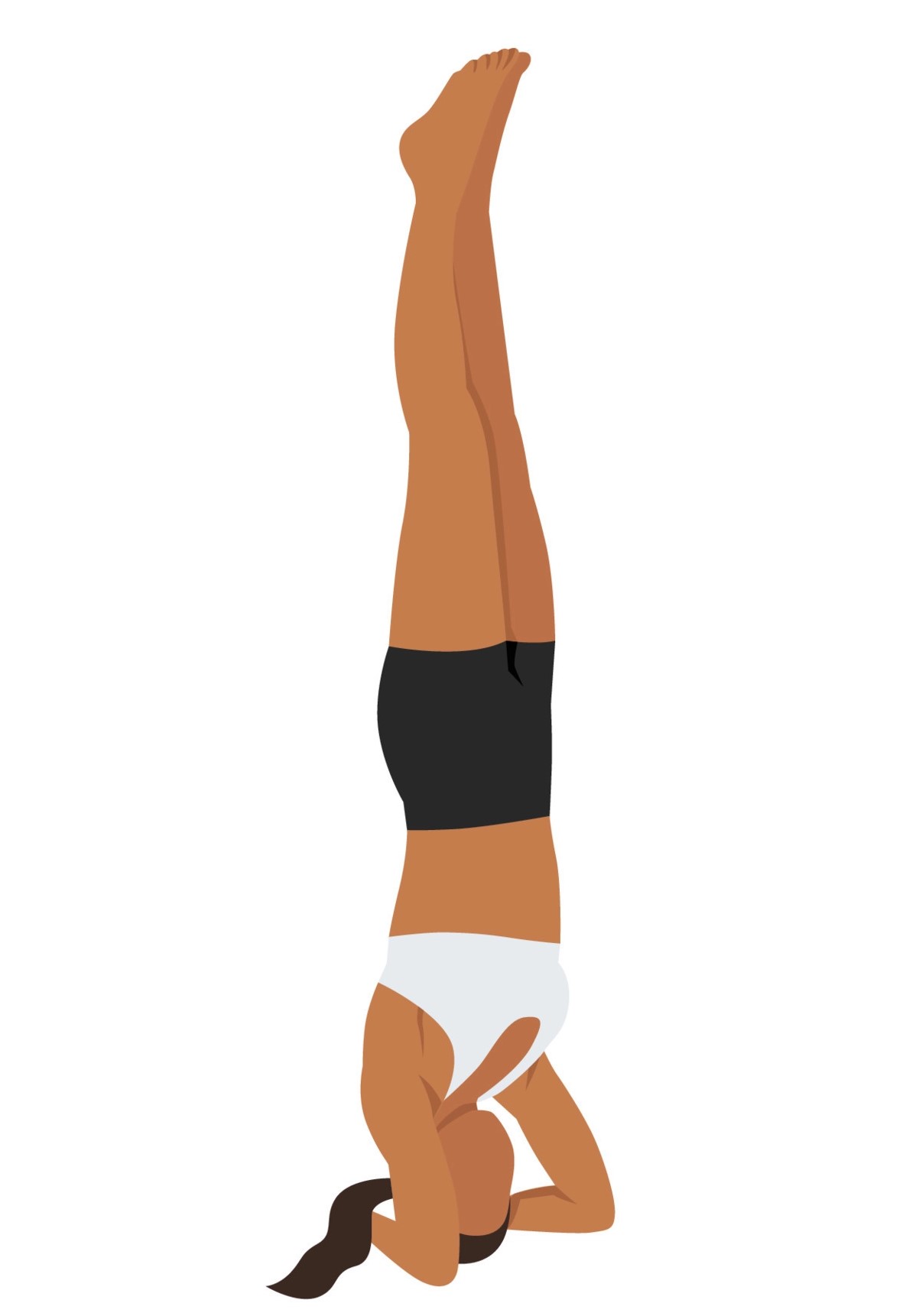} \hfill
\includegraphics[width=0.15\linewidth, height= 1.5 cm]{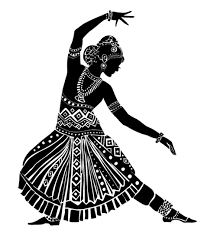}  \hfill
\includegraphics[width=0.15\linewidth, height=1.5 cm]{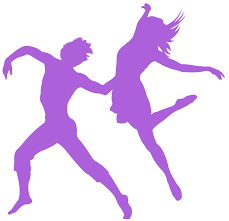} \hfill
}
\caption{Symbolic examples of fine-grained \textbf{S}ports, \textbf{Y}oga, and \textbf{D}ance (\textbf{SYD}) postures, represent complex body part movements and gestures, which are key challenges in posture recognition.}
\label{fig:syd1}
\end{figure}

\IEEEPARstart{H}{uman} body-pose recognition is a challenging problem in computer vision. It is widely used in various applications, such as  sports \cite{van2022deepsportradar, barbon2021sport,  hwang2017athlete,  dai2020human, thomas2017computer, zalluhoglu2020collective}, yoga \cite{ wu2022computer, verma2020yoga, yadav2022yognet, li2021and,  dittakavi2022pose,  yadav2022yogatube, chasmai2022view, wang2020swimming},  
dance \cite{samanta2014indian, kishore2018indian, tiwary2016classification, naik2020classification, bhuyan2021automated}, 
daily activity \cite{bera2022sr}, and others \cite{wu2022local, wang2020swimming,wang2018inertial}. Among these actions and postures, \textbf{S}ports, \textbf{Y}oga, and  \textbf{D}ance (SYD) are intrinsically very important physical activities to balance functionalities of various body parts,  well-being, etc. The SYD activities (Fig. \ref{fig:syd1}) are crucial to improve our quality of life (QoL) and  mitigating several diseases and mental health conditions, \textit{e.g.}, Parkinson’s disease, anxiety, sleeping disorder, etc. \cite{habib2022machine}.
Fine-grained image classification (FGIC) using SYD postures is  difficult due to huge intra-class variations and small inter-class differences among the sub-categories. SYD express our emotion, complex body movements, gestures, costumes, and diversity. 
Dance is a perceptual domain integrating  audio (music) and video (posture) in a synchronized manner to represent the underlying knowledge of a dance style \cite{mallik2011nrityakosha}. In this direction, Indian Classical Dance (ICD) and Yoga poses (aka $asana$)   effectively  represent their heritage and culture since ancient times \cite{samanta2012indian, bisht2017indian}. 
Also, hundreds of dancing themes (\textit{e.g.}, tribal, folk, etc.) are popular across the world. These diversities are promoted in the Intangible Cultural Heritage  of UNESCO, 
such as the  Lazgi (Khorazm region in Uzbekistan) \cite{skublewska2021methodology},  Thai (Thailand) Kolo (Serbia),  Lad’s (Romania), Bharatnatyam (India), and others. 

\vspace{-0.2cm}
To address the challenges in SYD recognition, existing methods have emphasized  hand-crafted features (\textit{e.g.}, bag-of-words,  haar wavelets, scale-invariant feature transform (SIFT), space-time interest points (STIP), moments, etc.) \cite{kumar2017indian, hassan2011annotating}, pose estimation from skeletal joints \cite{saha2013study,wang2018inertial}, motion analysis using optical flow information \cite{bisht2017indian}, and deep convolutional features \cite{  naik2020classification, parameshwaran2020unravelling, wu2022local}, shown in Table \ref{Study}. The pose estimation performance is remarkably improved using deep learning and related fusion-based techniques. Most of these methods are experimented with laboratory-simulated  dance videos collected from YouTube, or other web resources, summarized in Table \ref{Study}. The Leeds Sports Pose (LSP) dataset \cite{johnson2010clustered} represents 8 sports classes. The Yoga-82 \cite{verma2020yoga} comprised of 82 fine-grained yoga poses. { Another dataset containing 107 yoga classes, Yoga-107, is also experimented}. However, no image-based dataset with diverse variations of dance styles is publicly available. Our main motivation is to create  new image-based dance and sports action datasets. We have presented the Sports image dataset with 102 actions, and the Dance image dataset with 12 categories,  which are the first of their kind. A new deep model is devised to simulate experimental analysis on four SYD datasets in a weakly supervised manner. 

\begin{figure*}[h]
\centering
\subfloat{ 
\includegraphics[width=0.95\linewidth, ]{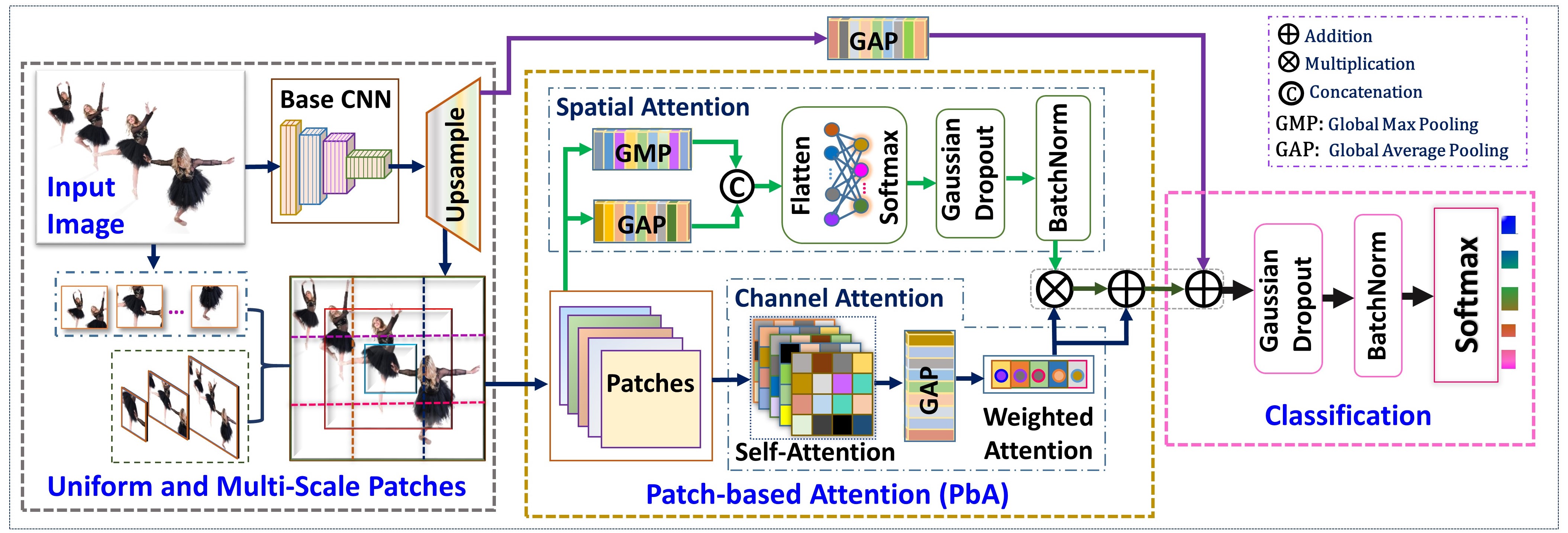}
}
\caption{{Proposed SYD-Net Model comprises  three key modules. a) Extraction of hybrid patches, consisting of same-size uniform patches  and hierarchical multi-scale regions. b) Patch-based attention module, consisting of the self-attention feature maps and further refined with the weighted-attention to define channel-attention. It is modulated with a spatial attention module to define an attentional feature map. c) Finally, the compact feature vector is regularized with the Gaussian dropout and batch normalization prior to the softmax layer for classification.}}
\label{fig:Model2a}
\vspace{-0.3cm}
\end{figure*}

We have proposed a patch-based attention (PbA) module, namely SYD-Net (Fig. \ref{fig:Model2a}), that integrates  spatial attention and channel attention on top of standard backbone convolutional neural networks (CNNs). Our aim is to establish a semantic correlation among a set of uniform and multi-scale patches by focusing on the most relevant image regions for defining a comprehensive feature descriptor. SYD-Net is inspired by the  self-attention mechanism \cite{vaswani2017attention, bahdanau2014neural}, which  is an integral ingredient of numerous deep architectures in computer vision and natural language processing ubiquitously. 
Gaussian drop-out \cite{reygaussian} is adapted to hinder overfitting. Random region erasing \cite{zhong2020random} data augmentation (Fig. \ref{fig:patches}) produces on-the-fly data diversity for effective  training of SYD-Net. The major contributions of this work are:

\begin{itemize}
    \item A patch-based attention mechanism that  summarizes the discriminativeness of partial feature descriptors for fine-grained sports, yoga, and dance postures recognition.
    \item A new image dataset with 102 sport actions and another dataset representing 12 dance styles are proposed for  posture classification avoiding part-based/skeletal-joint/bounding-box information. 
    {\item Extensive experiments are conducted using five backbone CNN architectures on four SYD datasets in a weakly supervised manner.}
    {\item The proposed SYD-Net approach achieves state-of-the-art accuracy on the Yoga-82 and Yoga-107 datasets.} 
\end{itemize}

The rest of this paper is organized as follows: Section \ref{Rwrk}
summarizes related works on SYD poses. Section \ref{method} describes the proposed method.  Section \ref{datades} describes the datasets briefly. Section \ref{expmnt} demonstrates the  experimental results of the ablation study. The conclusion is presented in Section \ref{con}.

\begin{table*}
\caption{Summary of a Few Existing  Sports, Yoga, and Dance Recognition Image/Video Datasets with Related  Descriptions. } 
\begin{center}
\begin{tabular}{|p{1.3 cm}|p{6 cm}|p {5.1 cm}|p {3.8 cm}|}
\hline
Ref, year & Method & Input data & Dataset name and pose/actions  \\
\hline
\cite{johnson2010clustered}, 2010  &  appearance and pose-based pictorial structure model. & 1000 training and 1000 testing images & Sports 8: Soccer, Tennis, etc.  
\\   
\hline
\cite{lan2011discriminative}, 2011 & global scene model with a
figure-centric  visual word representation  & 150 clips (mean 6.39s) at 10fps   & UCF Sports: kicking, lifting, riding horse, running,  etc. \\
\hline 
 \cite{karpathy2014large}, 2014 & multiple schemes to learn spatio temporal features & 1 million YouTube videos, avg 5.36 min clip. & 487 classes: cricket, racing, etc. \\
\hline
\cite{maddala2019yoganet}, 2019 &   spatio-temporal color-coded image called,  joint angular displacement map (JADM).  & 16,800 3D yoga videos  with 400 videos per pose, collected using a mocap setup.   & Yoga 42: standing postures, katichakrasana, etc.\\
\hline
\cite{verma2020yoga}, 2020 &  classification using CNNs in a hierarchical structure  & 28,478 yoga-pose images are
collected from various search engines.
& Yoga 82: Handstand, Plank, Side reclining, Shoulderstand, etc. \\
\hline
\cite{wu2022computer}, 2022 &  contrastive skeleton feature
representations and  extracts 33 keypoints using Mediapipe   &
 contains 1931 images collected from Kaggle &
Yoga 45: balancing, sitting, upward bow pose,  etc. \\
\hline 
\cite{samanta2013novel}, 2013 &
space-time interest point  descriptors computed  from each frame and classified  by a non-linear SVM. 
&  330 video clips of 87  dancers at 30 fps, and  collected from stage performance. 
& ICD  -  6 styles: Bharatanatyam, Kathak,  Odissi,
 etc.  \\

\hline
\cite{han2016dancelets}, 2017 & mid-level action representation using dancelets for dance-based video recommendation. 
&   420 videos collected from YouTube and Youku websites.  &   HIT Dances 6 styles: Ballet, Hip-hop and 4 Chinese folks. \\
\hline
\cite{castro2018let}, 2018 &integrates optical flow and
 motion data.  Fusion-based multi-stream 3D temporal CNNs were tested. & 1000 videos of 10 sec. at 30 fps, a total of 300000 frames.   &  Let’s Dance 10 classes: Ballet,  Foxtrot, Latin, Tango, etc.\\
\hline
 \cite{hu2021unsupervised}, 2021 & 
 Estimates 2D  pose sequences, and tracks dancers. Simultaneously estimates corresponding 3D poses.  
 &   1143 video clips of 9 genres,
 154 movement types of 16 body parts. & University of Illinois Dance - 9  types: Ballet, Tango, etc.  
 \\  
\hline
\cite{shailesh2022understanding}, 2022 & understanding dance semantics by spatio temporal dynamics  using keypoints (OpenPose) and GRU. & 300 HD dance videos, collected from 6 performers in indoor and outdoor, and YouTube.
& 7 types:  firing arrows, dance of peacock, playing flute, etc. 
\\
\hline
{Ours, 2023} & {Multi-scale patch-based attention mechanism} & {5967 image samples collected from Kaggle} & {Yoga 107: tulasana, lolasana, etc.} \\

\hline 
\end{tabular}
\label{Study}
\end{center}
\vspace{ -0.5 cm}
\end{table*}

\section{Related Works} \label{Rwrk}
Posture recognition methods can be generalized  into two broader streams: a) hand-crafted, 
and b) deep-learning methods including, attention-based works. Some SYD  pose recognition methods are summarized in Table \ref{Study},  and described  next.

\vspace{-0.2cm}
\subsection{{Study on Sports Actions}}
Sports activities are primarily recognized using deep architectures. Various sports technologies and related datasets are studied and analyzed  the role of computer vision technologies in sports  (\textit{e.g.}, player and ball tracking) in \cite{thomas2017computer}. The pictorial structure model (PSM) with clusters of  partial pose descriptors from sports images is presented  \cite{johnson2011learning}. The dataset is more challenging than its earlier version, called Leeds Sports Pose (LSP) dataset \cite{johnson2010clustered}. The collective sports video dataset represents a  multi-task recognition of both 5 collective activities and  11 sports categories \cite{zalluhoglu2020collective}.  Deep learning methods using CNNs and long short-term memory (LSTM) are tested for benchmark analysis.  A two-stream attention model using LSTM is described for action recognition \cite{dai2020human}. Freestyle wrestling actions are recognized from videos using a histogram of graph nodes \cite{mottaghi2020action}. A multi-labels DeepSport dataset is presented for automated sport understanding using videos captured at multiple views of a basketball game  \cite{van2022deepsportradar}. A CNN combining  global regression and local information refinement modules for sports-pose estimation  using 2D images is presented  \cite{hwang2017athlete}. Swimming motion analysis is presented  \cite{wang2020swimming}. Figure Skating Dataset with 10 sports actions (FSD-10) is introduced  for fine-grained content analysis using a key-frame-based temporal segment network \cite{liu2020fsd}. Most of the existing works on sports analysis are based on video datasets.
\vspace{-0.2 cm}
\subsection{Study on Yoga Postures}
Mainly, three types of intelligent approaches for yoga pose analysis   have been developed: (a) wearable device,  (b) Kinect, and (c) computer vision. A  hybrid multi-modal and body multi-positional system for recognizing 21 complex human activities using wearable devices is developed \cite{bharti2018human}. The CNNs  recognize yoga  poses from 3D motion capture data by integrating a joint angular displacement map (JADM) comprising  39 joints of yoga action skeletons  \cite{maddala2019yoganet}. 
A yoga pose grading approach is described  using contrastive skeleton feature representations  \cite{wu2022computer}. 
Many approaches have used OpenPose to estimate the keypoints/joints in developing Yoga pose recognition and mobile applications for self-assessment and yoga assistance \cite{yadav2019real,  lin2021openpose, bhambure2021yog}. 
A yoga self-coaching system  using an interactive display in real time is developed  to avoid incorrect postures \cite{long2021development}. It classifies 14 yoga postures based on transfer learning. 
The fitness actions of 28  poses are classified into   three categories of  exercises \cite{li2021and}.
Likewise, 88 videos are used for classifying 6 Yoga poses \cite{yadav2019real}. Yoga-82 \cite{verma2020yoga} has introduced a new dataset containing almost 28.5k images of 82 fine-grained yoga poses, and illustrated in Fig. \ref{fig:yoga82_samples}. This dataset is tested in our study for further improvement. 
\vspace{-0.2cm}
\subsection{Study on Dance Postures}
Traditional hand-crafted feature-based and deep learning approaches are developed for dance posture recognition. The bag-of-words method is applied to recognize five Greek dance styles from videos \cite{kapsouras2013feature}. The space-time interest point (STIP) detection and their description from videos using a 3D facet model are presented  \cite{samanta2014space}. 
A spatio temporal Laban feature descriptor (STLF)  from YouTube videos is described  \cite{dewan2018spatio}. 
Using a Kinect sensor, 3D skeletal information  of 25 leg postures from five dancers representing $Odissi$ dance  has been collected, and a similarity function is used for recognition \cite{saha2013study}.  
Multiple kernel learning using a directed acyclic graph is presented  \cite{hassan2011annotating}. 
With  music, $Kathakali$ demonstrates complex  hand gestures, body movements, and facial expressions. The dataset contains 654 images representing  24 mudras of $Kathakali$ and was tested  using CNN  \cite{bhavanam2020classification}.  
$Bharatanatyam$ posture recognition is tested on audio  and video  data using the Gaussian mixture model (GMM), support vector machine (SVM), and CNN \cite{mallick2019posture}. 
Dance semantics  understanding from videos by deep pose estimation (based on OpenPose) coupled with a gated recurrent unit (GRU) is presented \cite{shailesh2022understanding}.
The  Inception-v3 features, 3D CNN features, and pose signature based on AlphaPose estimation  are combined and  fed into an LSTM for building spatio temporal relationships for ICD classification \cite{kaushik2018nrityantar}.
A model based  on ResNet-50 recognized eight ICD \cite{jain2021enhanced}.
An Uzbek national dance,  $Lazgi$  classification and recognition using an optical motion capture system are explored   \cite{skublewska2021methodology}.
According to our study, image datasets representing  various sport actions and dance styles are unavailable for weakly supervised pose estimation. This work presents two new image datasets for sport and dance actions recognition. 

\begin{figure*}[h]
\centering
\subfloat[Fixed RGB=127]{ 
\includegraphics[width=0.09\linewidth, height= 1.50 cm]{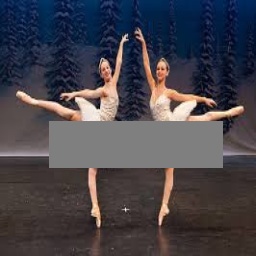}
\includegraphics[width=0.09\linewidth, height= 1.50 cm]{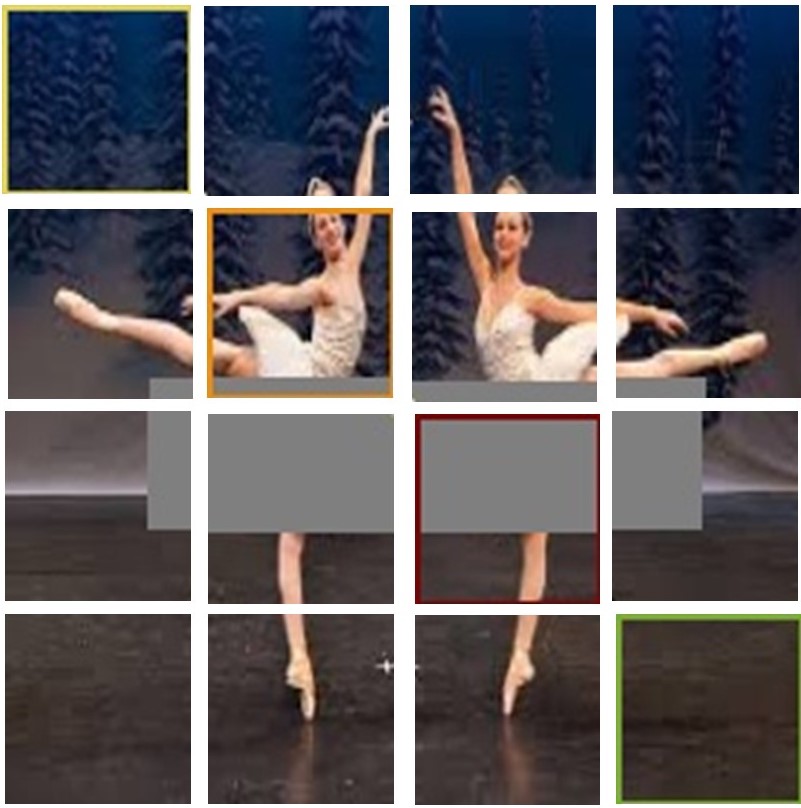}
\includegraphics[width=0.09\linewidth, height= 1.50 cm]{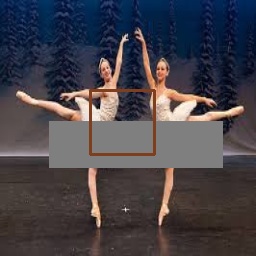}
\includegraphics[width=0.09\linewidth, height= 1.50 cm]{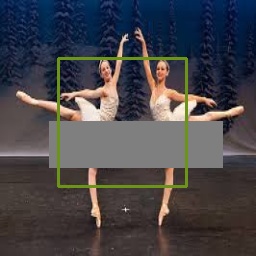}
\includegraphics[width=0.09\linewidth, height= 1.50 cm]{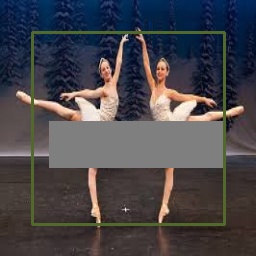}
}
\subfloat[Random RGB ]{ 
\includegraphics[width=0.09\linewidth, height= 1.50 cm]{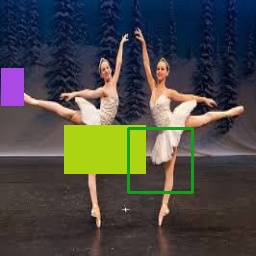}
\includegraphics[width=0.09\linewidth, height= 1.50 cm]{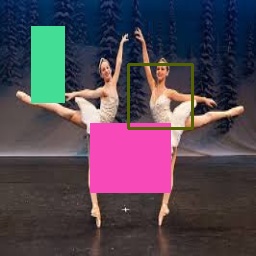}
\includegraphics[width=0.09\linewidth, height= 1.50 cm]{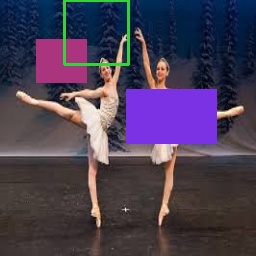}
\includegraphics[width=0.09\linewidth, height= 1.50 cm]{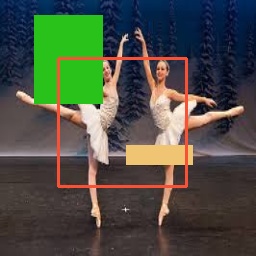}
\includegraphics[width=0.09\linewidth, height= 1.50 cm]{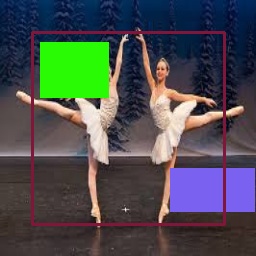}
}
\caption{Patches with random erasing image augmentation. (a) Erased  full image with fixed RGB=127; 4$\times$4 uniform, and  multi-scale patches, enclosed by rectangles. (b) Two random regions erased with random RGB colors on both types of patches.}
\label{fig:patches}
\vspace{-0.4 cm}
\end{figure*}

\section{Proposed Method} \label{method}
{Sports, Yoga, and Dance (SYD) involves complex  movements of body parts and subtle variations in expression and gesture, \textit{e.g.}, yoga and dance poses. Human-object interactions could also be involved, \textit{e.g.}, players with a football in sports. Various pose-estimators, object detectors, skeletal joints, and body parts are often used to solve the problem. The recognition task becomes more challenging when multiple persons are involved in an activity. In some cases, extraction and localization of body keypoints of multiple persons from a still image could be burdensome, and difficult to formulate an appearance-based model. Sometimes, a global descriptor overlooks finer details, which are essential for FGIC \cite{bera2021attend}. Our intuitive  approach is that region-based partial feature description could be an alternative solution in capturing finer details of SYD poses for classification. Our target is to devise an end-to-end trainable deep architecture to classify these complex fine-grained human postures avoiding any bounding-box annotation, object/pose detector, and body keypoints/joints, commonly used in existing works. Moreover, the random erasing technique with conventional data augmentation is followed for additional benefits to ease overfitting and  overall performance gain.} The proposed SYD-Net, conceptualized in Fig. \ref{fig:Model2a}, is divided into three parts: a) computing a set of non-overlapped patches with fixed-size and multi-scale region proposals, b) patch-based attention module, and c) classification. Functionalities of all modules are described next. 

\subsection{Uniform and Multi-Scale Patch Proposals} 
In an image, contextual information and associated object(s) provide a vital cue to understand human activity, evident in various sports (Fig. \ref{fig:s102}). Here, a patch-based approach is devised to learn  overall semantics and contexts from various image parts at multiple scales (Fig. \ref{fig:patches}). We aim to integrate detailed information from several smaller non-overlapping image patches into a comprehensive feature descriptor. Moreover,  hierarchical regions establish a semantic correlation and contextual representation among the feature maps. The uniform patches focus on finer details in each small  region. Whereas larger multi-scale patches summarize overall feature representation  holistically. Thus, combining  these two key aspects through an attention mechanism improves overall  efficiency for subtle discrimination in  fine-grained SYD postures. 

Let a color input image $I_l$ $\in$ $\mathbb{R}^{h\times w\times 3}$ is fed into a backbone CNN, such as MobileNet-v2 with class-label $l$. A backbone network $\mathcal{N}$  extracts  high-level feature maps $\textbf{F}$ $\in$ $\mathbb{R}^{h\times w\times c}$ where $h$, $w$, and $c$ denote the height, width, and channels, respectively. The input image $I_l$ is divided into a set ($\textit{D}$) of  non-overlapping  uniform patch proposals. The  resulting number of small regions is $e={(h\times w)}/a^2$, where $a\times a$ is the spatial size of a rectangular patch $d$. Set $\textit{D}=\{d_1,d_2,..., d_e | I_l\}$ consists of $e$ parts. A small patch $d_i$ is defined with its spatial dimension $ p_i=[x_i, y_i, \Delta w, \Delta h]$, and here $\Delta w=\Delta h=a$ is  uniformly the same for all patches. 
Each patch is denoted as $d_i$=$[\textbf{F}_i, p_i]$, where $\textbf{F}_i$ is the feature map of patch $p_i$. In addition,  multi-scale patches are defined to capture complementary information hierarchically, where the patch sizes are progressively increasing. It can be defined as 
$p_i=[x_i, y_i, \Delta w_i, \Delta h_i]$ and $p_j=[x_j, y_j, \Delta w_j, \Delta h_j]$ such that $\Delta w_i > \Delta  w_j$ and $\Delta h_i > \Delta  h_j$, where $p_i > p_j$ regarding the spatial dimension of $p_i$ and $p_j$ patches, respectively. Finally, a collection of all $n$ patches (\textit{i.e.}, uniform and multi-scale) is denoted as $P=\big\{p_i\big\}_{i=1}^{i=n}$, and corresponding feature map is $\textbf{F}=\big\{\textbf{F}_i\big\}_{i=1}^{i=n}$ $\in$ $\mathbb{R}^{n\times (h\times w\times c)}$. 
The feature map of each patch is determined through a mapping between the correspondence of smaller regions within the actual high-level output feature maps, extracted using a base network $\mathcal{N}$.  Firstly, $\textbf{F}$ is upsampled to a higher resolution $k(h\times w)$ for this intent. Then, bilinear interpolation is applied for pooling features from every patch. The  upsampling  is regarded as a mapping $m:$ ${\textbf{F}} \rightarrow {\textbf{F}}\in \mathbb{R}^{k(h\times w)\times c}$, where actual spatial size $(h$$\times$$w)$ is scaled up by $k$ times before pooling. Though $P$ represents patches of various sizes, bilinear pooling renders the feature vectors of the same sizes for all patches, which are kept the same as the output dimension of base CNNs, \textit{i.e.}, $\textbf{F}\in$ $\mathbb{R}^{(h\times w\times c)}$, and   denoted as $\textbf{F}$. %

\subsection{Patch-based Attention (PbA) mechanism}\label{sec:attn}
Attentional feature description is proliferated ubiquitously in image recognition and others to improve performance. Here, attention is performed in two paths, patch-based channel attention and spatial attention, which are finally integrated together. It fuses both to summarize essential features by exploring \textit {where} to focus and \textit {what} to emphasize simultaneously in the feature maps $\textbf{F}$.
Self-attention acts across the channel-based feature maps of all patches to capture channel-wise relationships. It relates inter-channel feature interactions among patches and estimates their relevance correspondingly. Cross-channel attention  investigates the importance of  feature maps  (\textit{what}) to enhance learning capability. On the contrary, spatial attention  explores neighborhood structural interpretation for producing a spatial attentional mask (\textit{where}) for further refinement of  aggregated feature summarization. These dual-attention pathways empower significantly and act complementarily  to render a global information for distinguishing  subtle variations in SYD recognition. 

\subsubsection{Channel Attention (CA)}
Channel attention is adapted from self-attention mechanism that tackles long-range dependency by generating  a context vector based on the weighted sum of  feature space \cite{vaswani2017attention}, \cite{bahdanau2014neural}. 
In self-attention, the {query} \textbf{Q}, {key} \textbf{K}, and {value} \textbf{V} are learned from the same input feature vector \textbf{F}. 
The attention-weight matrix is a dot product of \textbf{Q} and \textbf{K}, multiplied by \textbf{V} to generate an attention-focused feature map. We have applied $[Q, K, V]$ attention principle for a patch $p_i$ and its neighbors  $p_j$ patches ($i\neq j$). We aim to generate an attentional feature descriptor, \textit{i.e.}, {value} \textbf{V} that  focuses on the relevant and discriminative regions. 
$\textbf{F}_i$ and $\textbf{F}_{j}$ are  high-level feature vectors computed from $p_i$ and $p_j$ patches, respectively. The attentional feature map is given as
\begin{equation}
\vspace{-0.2 cm}
\centering
\begin{split}
    \psi_{i,j} &={tanh}({\mathbf{W}_\psi \textbf{F}_i} + {\mathbf{W}_{\psi'}\textbf{F}_{j}}+\mathbf{b}_{\psi}), \\
        \vartheta_{i,j}&=\sigma\left(\mathbf{W}_\vartheta \psi_{i,j} +  \mathbf{b}_{\vartheta} \right) \\
\end{split}
\vspace{-0.2 cm}
\end{equation}
where $\textbf{W}_{\psi}$ and $\textbf{W}_{\psi'}$ are the weight matrices to compute attention vectors using $p_i$ and $p_j$ patches, respectively;  $\textbf{W}_{\vartheta}$ is their nonlinear combination; $\mathbf{b}_{\psi}$ and $\mathbf{b}_{\vartheta}$ are the bias vectors, and $\sigma(.)$ is  element-wise  nonlinear activation. The next objective is to compute the importance of each $p_i$ through a weighted sum of  attention scores of all patches, given as
\begin{equation}
\centering
\vspace{-0.3 cm}
\begin{split}
    \delta_{i,j}={softmax}(\mathbf{W}_{\delta}\vartheta_{i,j}+\mathbf{b}_\delta), \text{ }
    \hat{\mathbf{F}}_i =  \sum_{j=1}^{{n}}&\delta_{i,j}\textbf{F}_{j} \\
    \end{split}
    \vspace{-0.3 cm}
\end{equation}
where $\textbf{W}_{\delta}$ is the weight matrix, and ${b}_\delta$ is the bias vector. The aggregated feature space is $\hat{\textbf{F}}_i$  which is summarized through a global average pooling (GAP) layer to generate ${\tilde{\mathbf{F}}}_i$ for all patches in $P$. The result ${\tilde{\mathbf{F}}}_i$ is passed through a \textit{softmax} layer for producing a weighted attention matrix $\phi_{i}$. Finally, their weighted sum  $\mathbf{F}_{CA}$ is considered as the output of the cross-channel attention (CA) mechanism, and is given as
\begin{equation}
\centering
\begin{split}
{	\tilde{\mathbf{F}}}_i = \mathcal{GAP}\left({\hat{\mathbf{F}}_i}\right) , \text{ }
{\mathbf{F}_{CA}} = \sum_{i=1}^{{n}}&\phi_{i}{	\tilde{\mathbf{F}}}_i  \\
\text{ where,} \hspace{2 mm} \phi_{i}={{softmax}}(\mathbf{W}_{\phi}{{	\tilde{\mathbf{F}}}_i}+\mathbf{b}_\phi) 
\end{split}
\end{equation}
\subsubsection{Spatial Attention (SA)}
Spatial attention captures the neighborhood information to calibrate feature representation by generating an attentional mask  for refining the global structural information. This mask builds spatial relationships by correlating  \textit{where to pay attention} in the feature space. Thus, it effectively localizes the most informative region(s) for global semantic  representation of SYD postures. 

No parameter optimization is required in the global average pooling (GAP), and it helps to avoid  overfitting \cite{lin2013network}. GAP sums out spatial information; thus, it is more robust to spatial translations of input. It can play as a structural regularizer in the network. Here, GAP is applied to refine spatial features  $\textbf{F}$ from  all patches $P.$ It downsamples the channel dimension precisely to $(h$$\times$$w$$\times$$1)$ by summarizing the mean features and generatig $\textbf{F}_{gap}$. Compared to GAP (.),  global max pooling (GMP) emphasizes the most important features from  cross-channels and generates an optimized feature vector $\textbf{F}_{gmp}$. A combination of both pooling  improves learning efficiency compared to any single pooling \cite{woo2018cbam}. The fused feature map  $\textbf{H}\in \mathbb{R}^{n(h \times w \times 2)}$ is defined as  
\begin{equation} \label{gap}
\centering
\begin{split}
\textbf{H}=concat\Bigl(\mathcal{GAP}\left( \textbf{F} \right); \mathcal{GMP}\left( \textbf{F} \right) \Bigl)  
\end{split}
\end{equation}
 where, the feature pooling is $\textbf{F} \rightarrow \textbf{F}_{gap}:\mathbb{R}^{n( h\times w \times 1)}$, and same for $\textbf{F}_{gmp}$.
 Next, a multi-layer perceptron (MLP) is applied to generate a spatial attention mask $\textbf{F}_{SA}$. A  MLP layer comprises a flatten, softmax, Gaussian dropout, and batch normalization layers. We aim to compute weighting factors based on the probabilities rendered by softmax activation. The probabilities are computed by a dense layer with the same size as base CNNs output channels $1\times c$. 
\begin{equation} \label{mlp}
\centering
\begin{split}
\textbf{F}_{SA}=  \mathcal{MLP} \big(\textbf{H} \big) \text{; }  { (softmax + \lambda_{GD+BN})\Rightarrow }\mathcal{ MLP}  \\  
\end{split}
\end{equation}
where $\lambda_{GD+BN}$ denotes a regularization ($\lambda$) process with a Gaussian dropout (GD) and batch normalization (BN) layers. The spatial attention mask is  $\textbf{F}_{SA} \in \mathbb{R}^{( 1\times c)}$.
This patch-level spatial attention ($\textbf{F}_{SA}$) mask is  multiplied element-wise with the weighted attention vector $\textbf{F}_{CA}$, rendered from channel attention method. It modulates overall feature representation and empowers discriminability  by capturing subtle details with focusing on global structural information, as essential for FGIC. A residual path is connected with $\textbf{F}_{CA}$  for smoother gradient flow in learning. Finally, a patch-based attention (PbA) feature vector $ \textbf{F}_{PbA} \in  \mathbb{R}^{(1 \times c)}$  is obtained.
\begin{equation} \label{PbA}
\centering
\begin{split}
\textbf{F}_{PbA}= \Bigl(\textbf{F}_{CA} \otimes \textbf{F}_{SA}  + \textbf{F}_{CA}\Bigl)   \\  
\end{split}
\end{equation}

\begin{figure} 
\centering
\subfloat{ 
\includegraphics[width=0.97\linewidth, ]{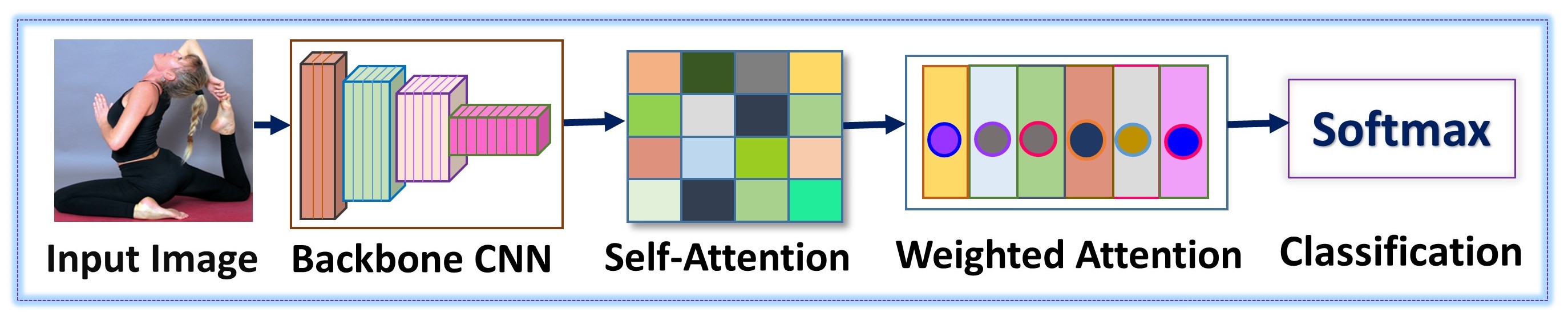}
}\vspace{- 0.3 cm}
\caption{{Baseline approach using attention  over CNN's output features.}}
\label{fig:Attn_Baseline}
\vspace{- 0.4 cm}
\end{figure}

\begin{figure*}[h]
\centering
\subfloat{ 
\includegraphics[width=0.12\linewidth, height= 1.8  cm] {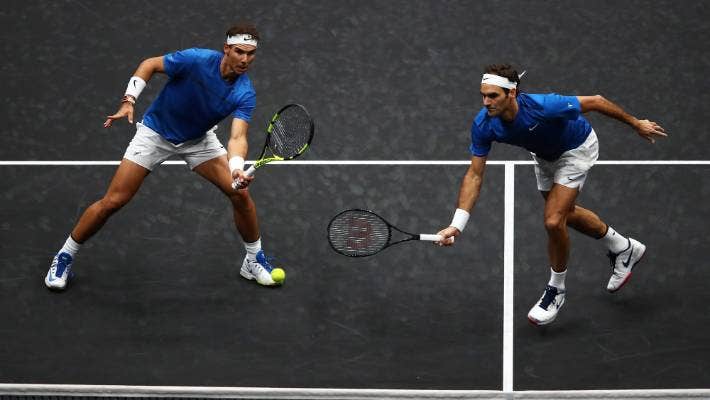} \hfill
\includegraphics[width=0.12\linewidth, height=1.8  cm] {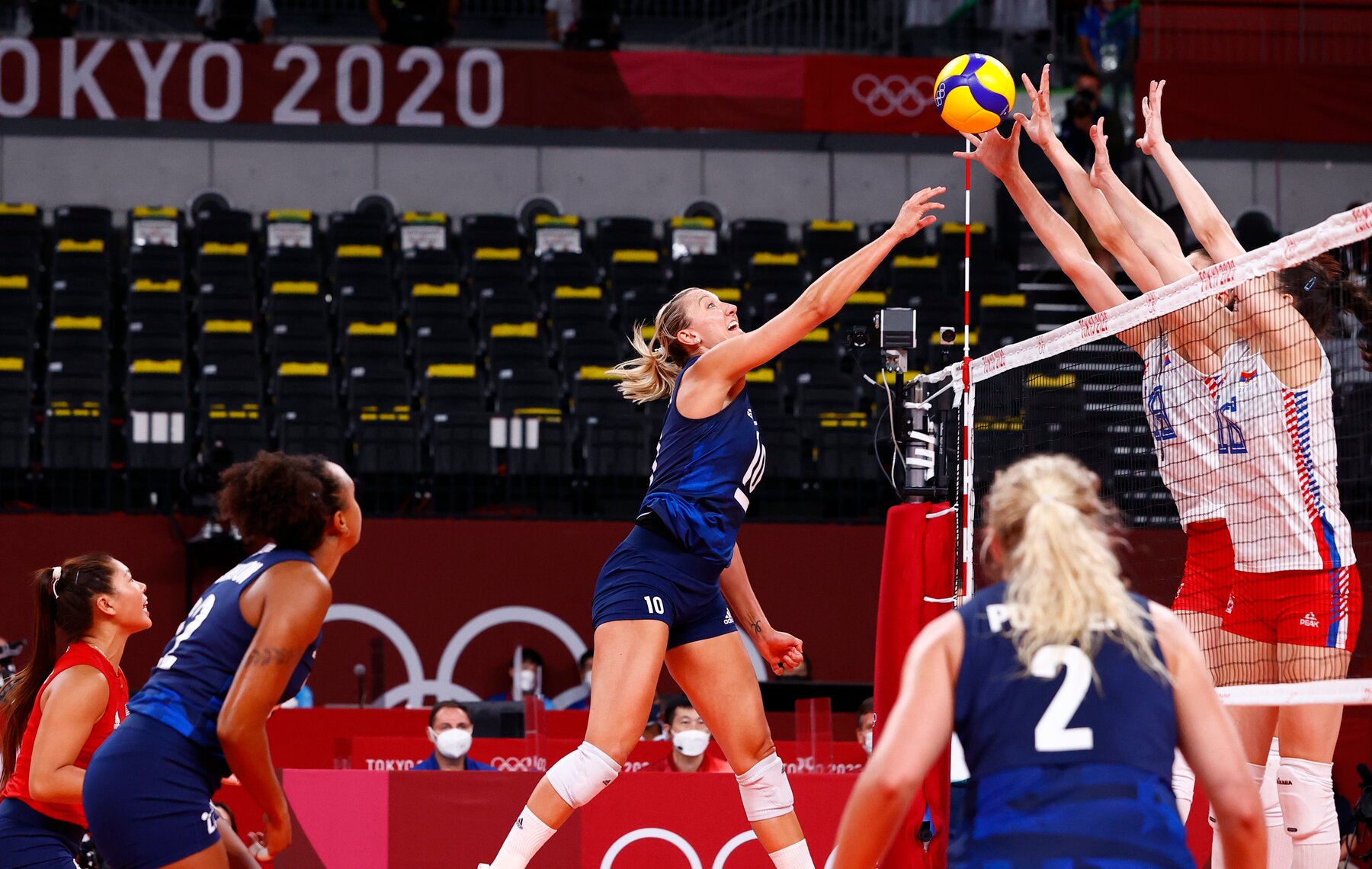} \hfill
\includegraphics[width=0.12\linewidth, height=1.8  cm] {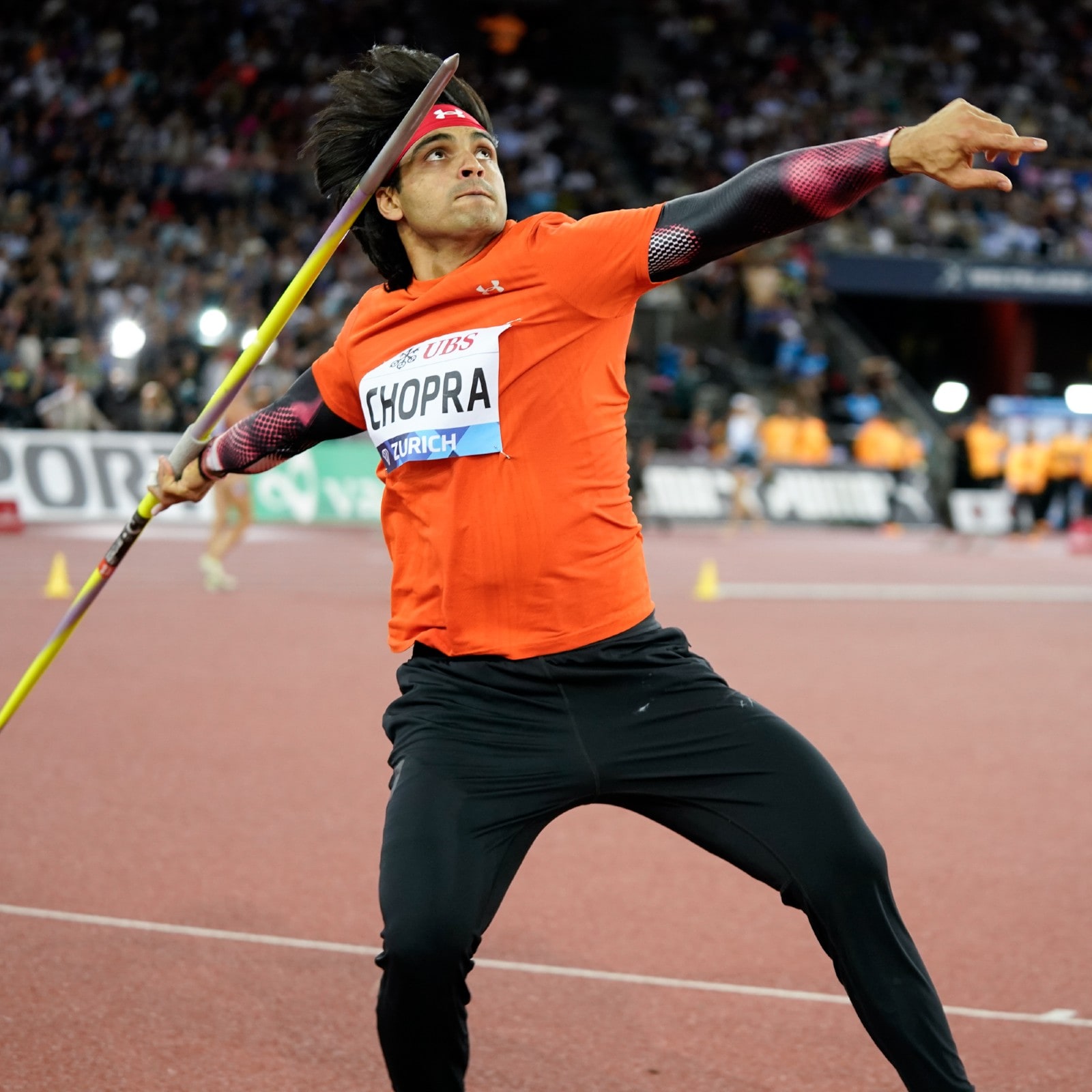} \hfill
\includegraphics[width=0.12\linewidth, height= 1.8  cm]{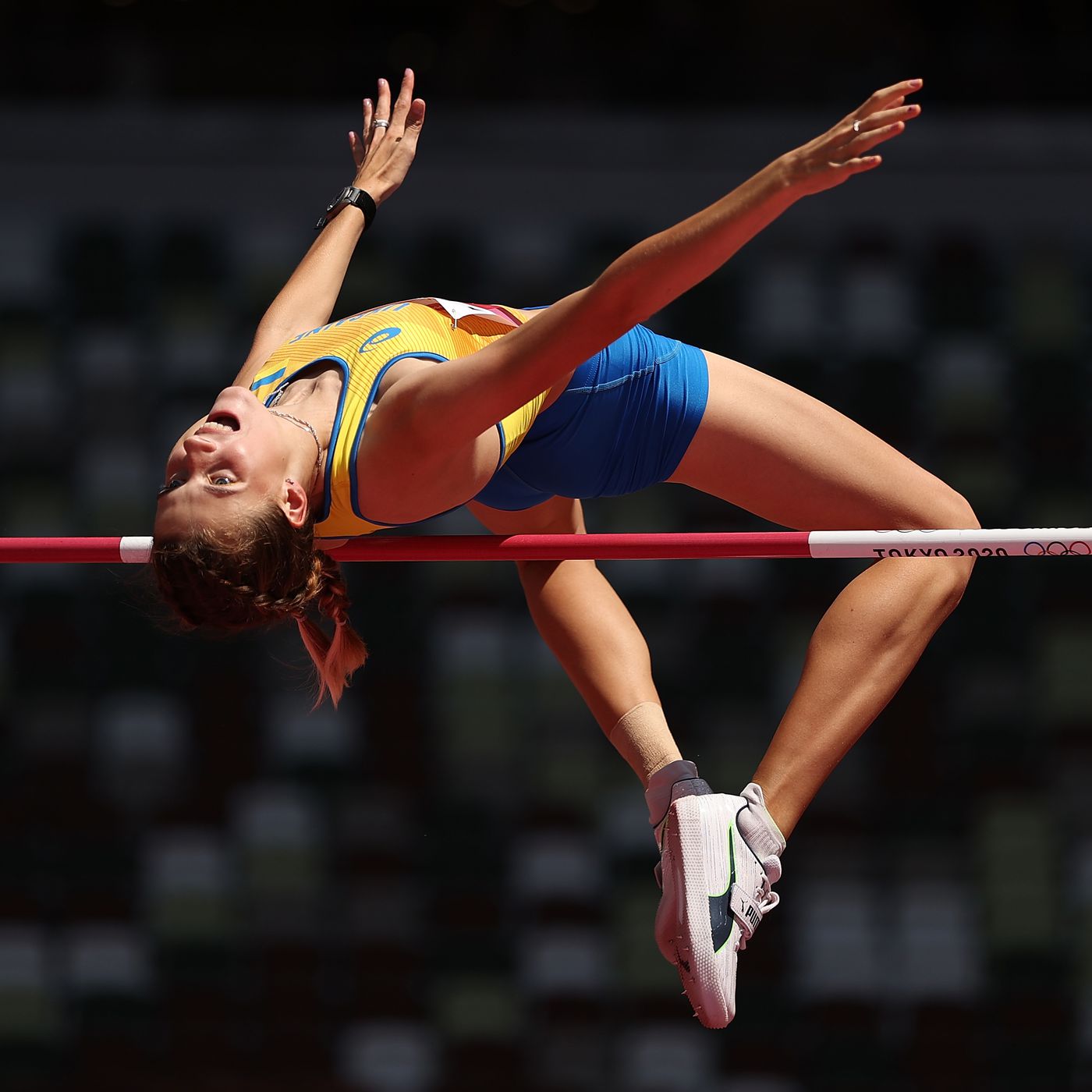} \hfill
\includegraphics[width=0.12\linewidth, height= 1.8  cm]{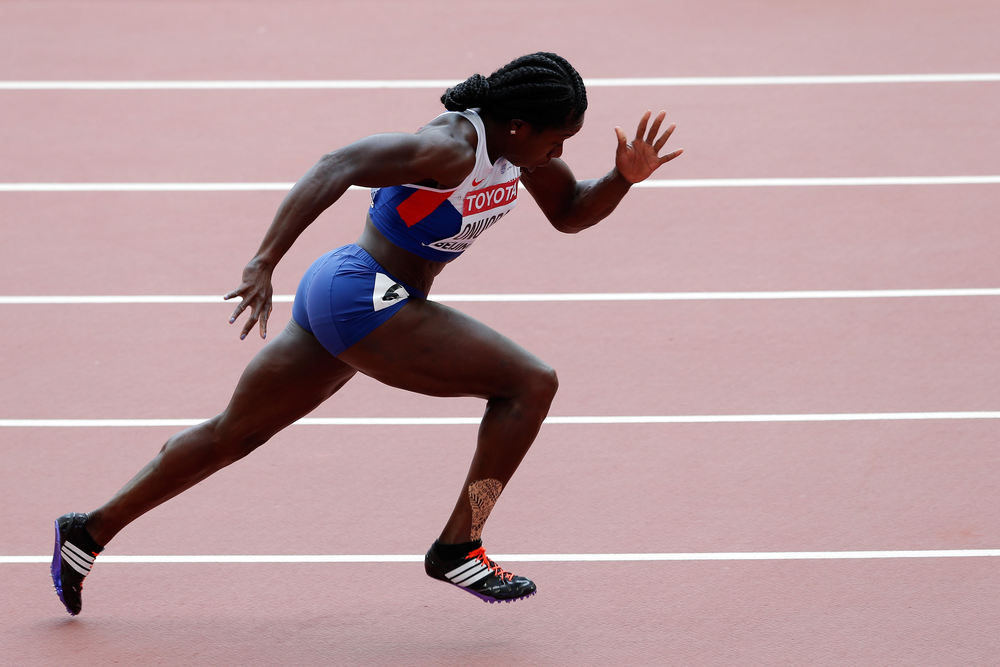}  \hfill
\includegraphics[width=0.12\linewidth, height= 1.8  cm] {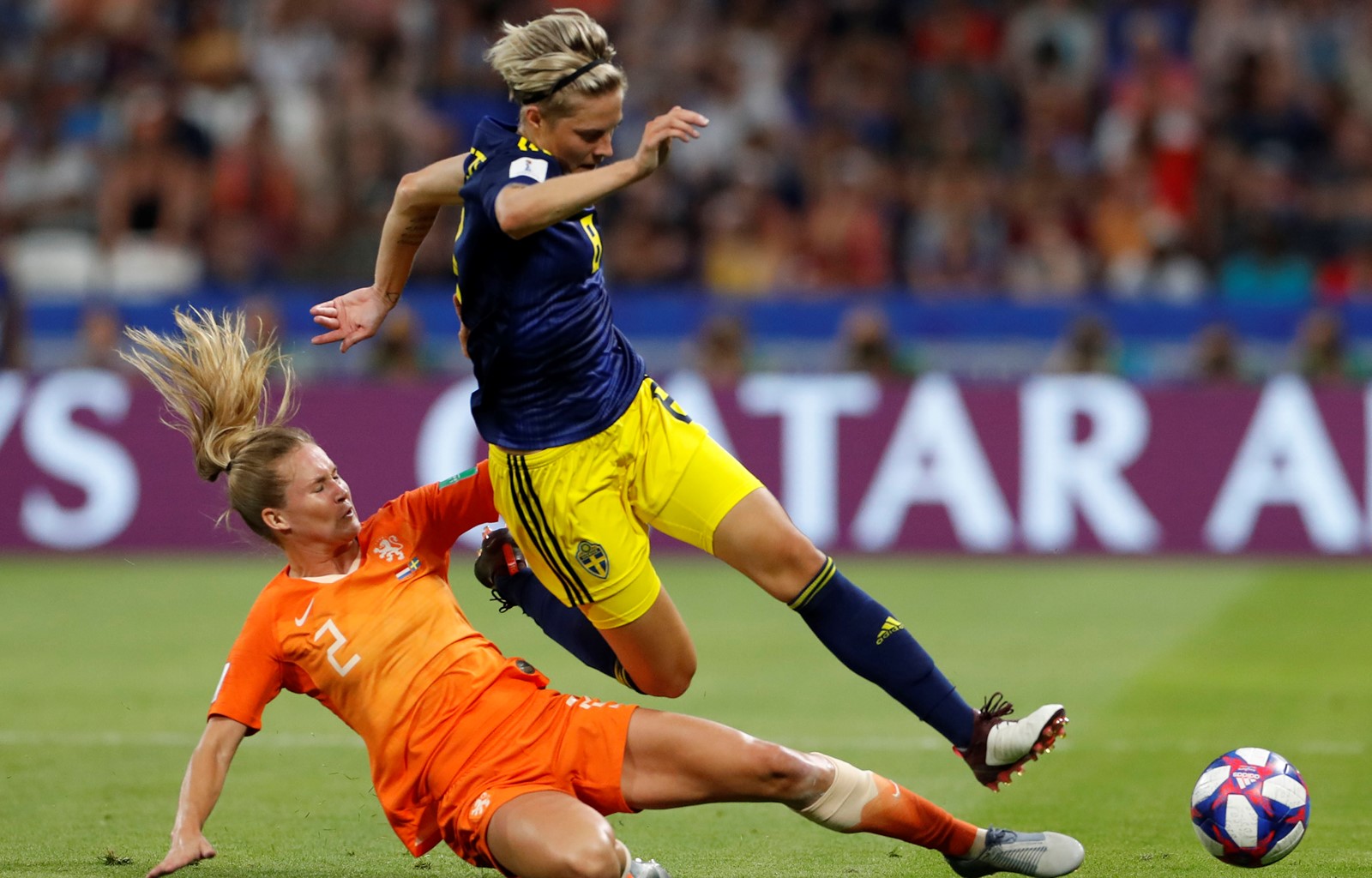} \hfill
\includegraphics[width=0.12\linewidth, height= 1.8  cm]{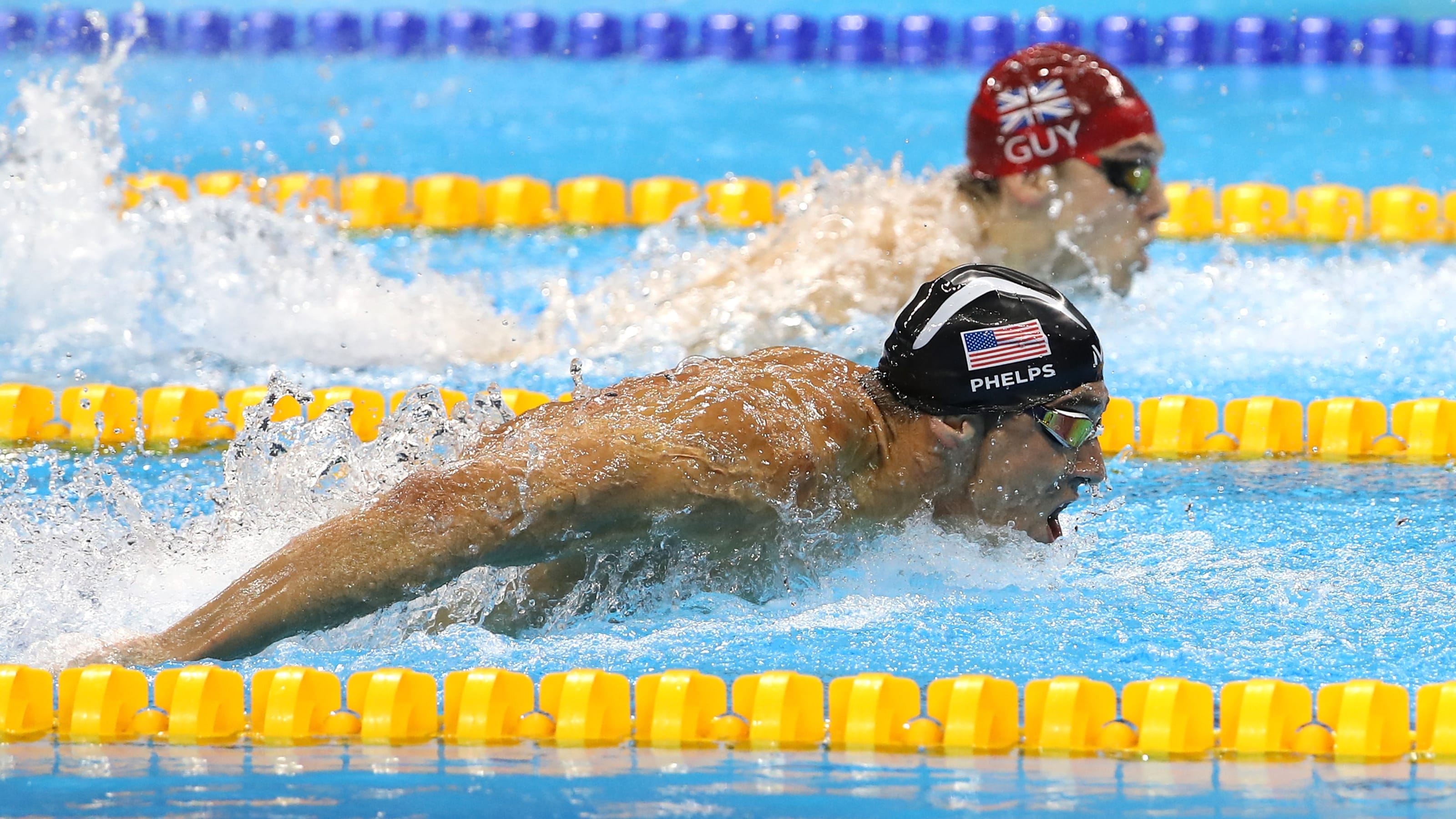} \hfill
\includegraphics[width=0.12\linewidth, height= 1.8  cm]{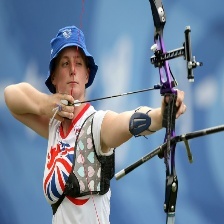} \hfill
}
\caption{Examples of diverse sport actions from the Sports-102 dataset.}
\label{fig:s102}
\end{figure*}

\begin{figure*}[h]
\centering
\includegraphics[width=0.95\linewidth, height=3 cm]{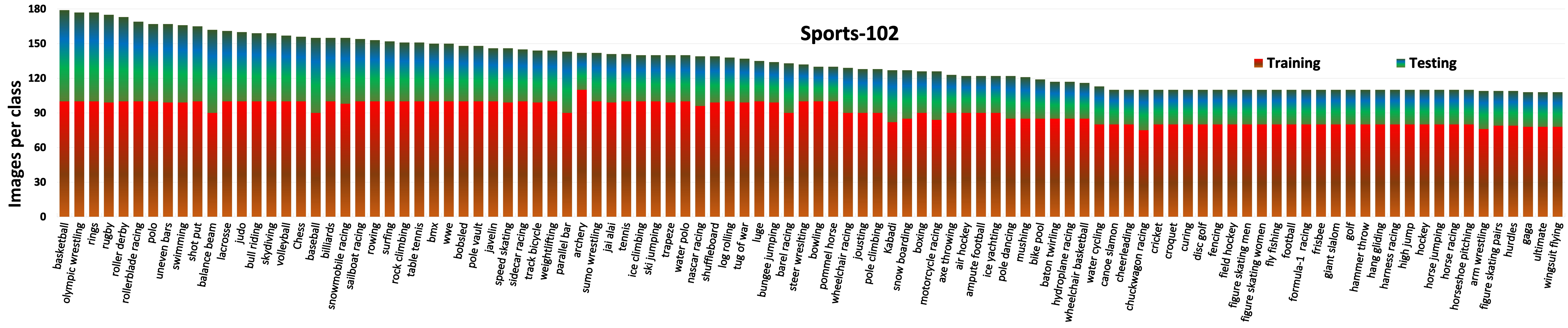} 
\caption{Class-wise train-test image distribution of Sports-102 dataset. Best view of the class labels in zoom.}
\label{fig:data_Sports}
\end{figure*}

\begin{figure*}[h]
\centering
\subfloat [Yoga-82 dataset]{ 
\includegraphics[width=0.12\linewidth, height= 1.8 cm]{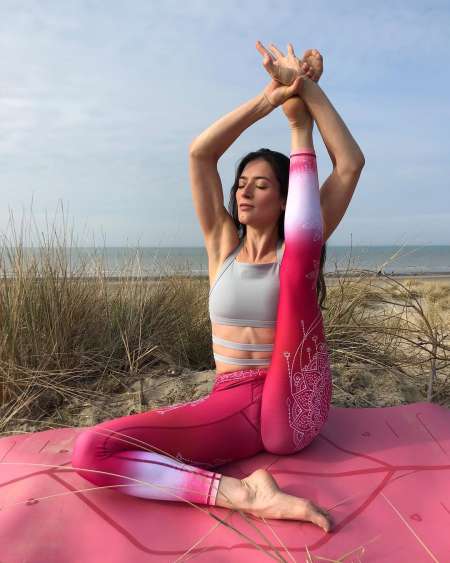}
\includegraphics[width=0.12\linewidth, height= 1.8cm]{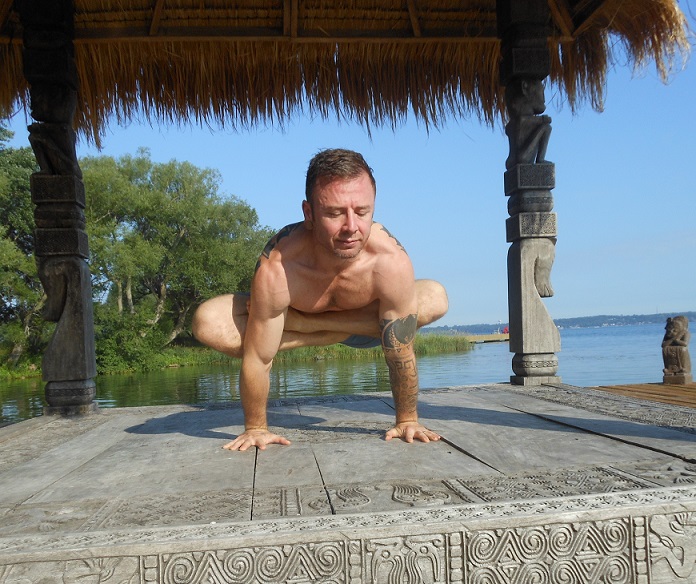}    
\includegraphics[width=0.12\linewidth, height= 1.8 cm]{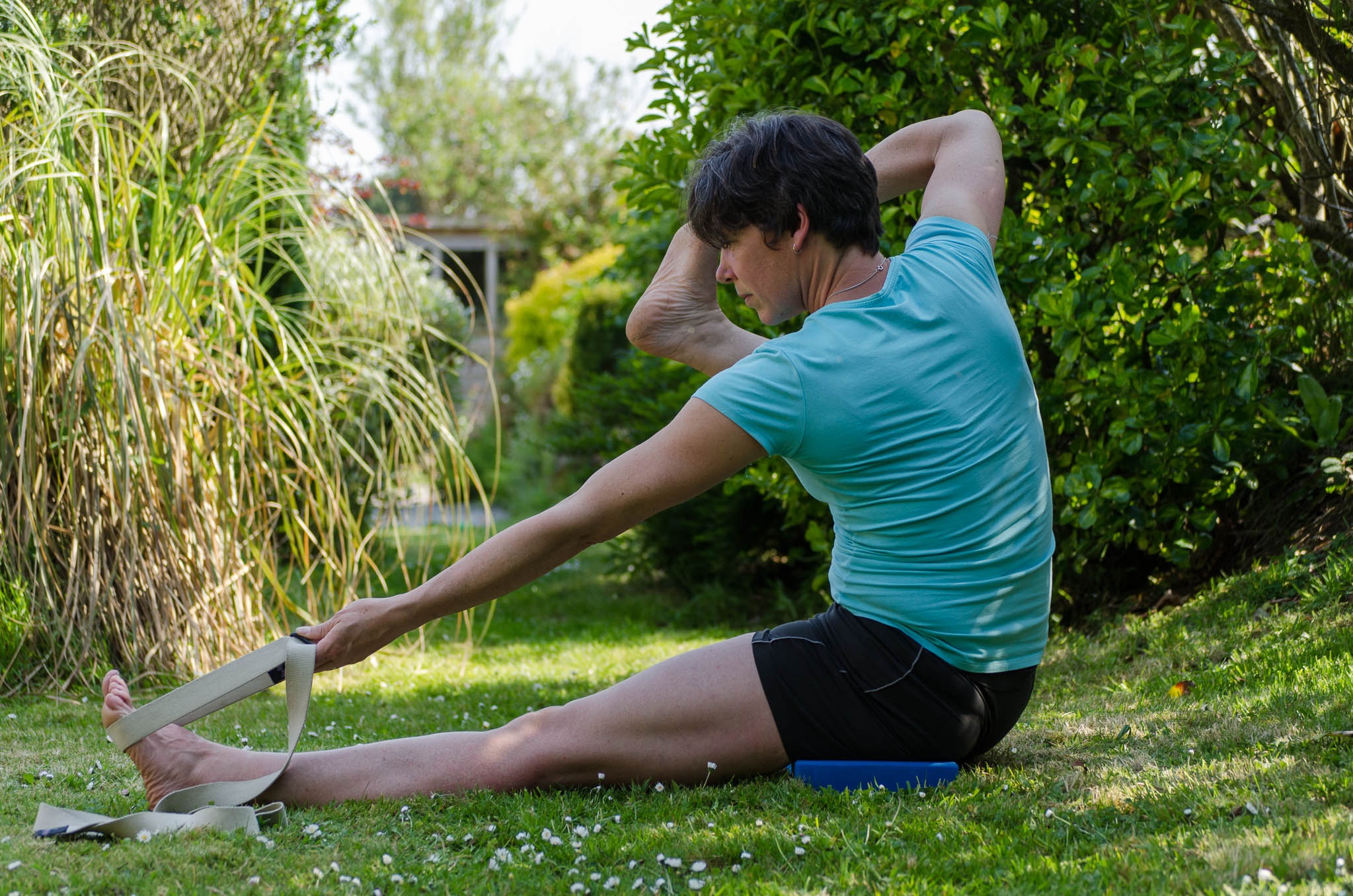}
\includegraphics[width=0.12\linewidth, height= 1.8 cm]{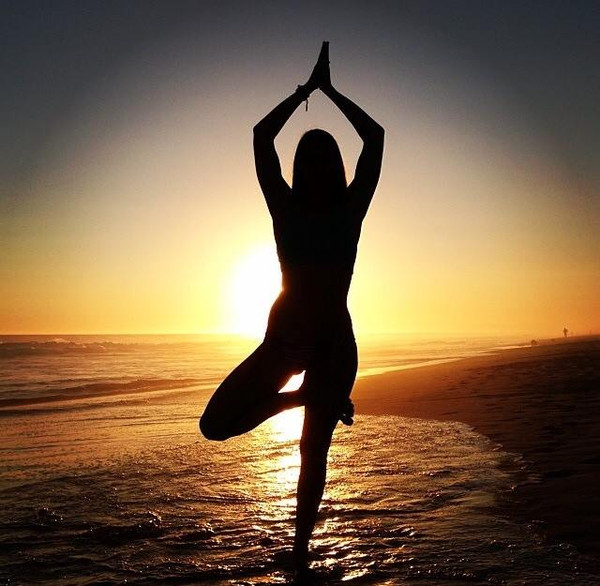}
}
\subfloat [{Yoga-107 dataset}]{ 
\includegraphics[width=0.12\linewidth, height= 1.8 cm]{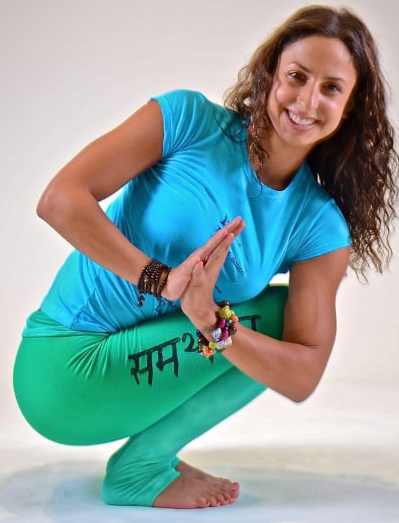}
\includegraphics[width=0.12\linewidth, height= 1.8 cm]{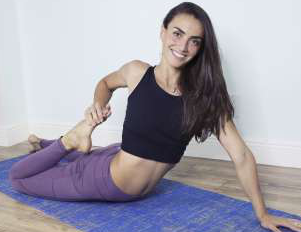}
\includegraphics[width=0.12\linewidth, height= 1.8 cm]{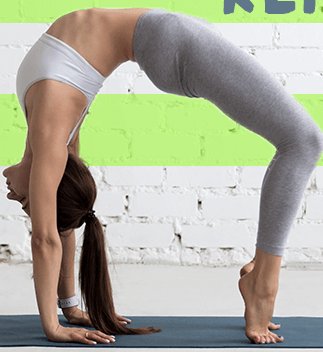}    
\includegraphics[width=0.12\linewidth, height= 1.8 cm]{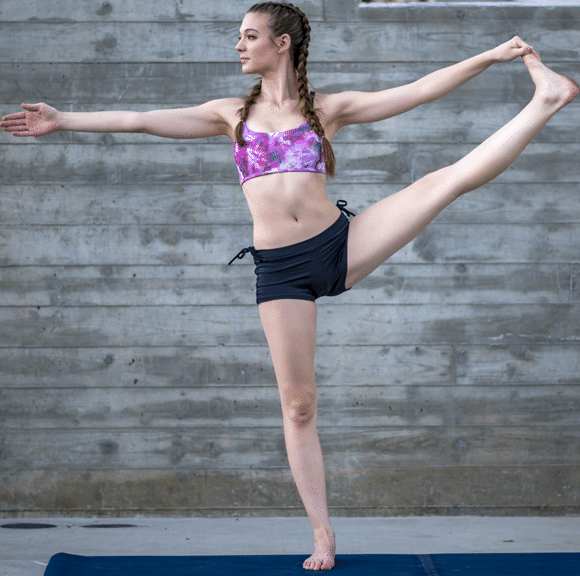}
}
\caption{Examples of various $asana$s from the Yoga-82 \cite{verma2020yoga} and {Yoga-107 datasets}. }
\label{fig:yoga82_samples} 
\end{figure*}  

\subsubsection{Classification}
The  upsampled feature $\textbf{F}$ is squeezed  by a GAP layer to produce a  vector  of $(1$$\times$$c)$ channels which is added with attentional feature map $\textbf{F}_{PbA}$. The final feature vector $\textbf{F}_{f}$ is regularized and  passed to a softmax layer to compute an output vector implying the  class probabilities.
\begin{equation}
\centering
\begin{split}
\textbf{F}_{f} =  \textbf{F}_{PbA}+ \mathcal{GAP}\left(\textbf{F} \right)  \text{ ; } Y_{pred}= softmax(\lambda(\textbf{F}_{f}) ) 
\end{split}
\end{equation}
Gaussian dropout (GD) \cite{reygaussian} and batch normalization (BN)  regularizers are applied to avoid overfitting issues and  denoted  as  $\lambda$. The GD can generalize learning tasks effectively than a simple dropout layer. Typically, GD  uses multiplicative noise, and the dropout rate $\phi$ maps to the noise standard deviation $\sigma_{noise}$. This hyperparameter is computed as $\sigma_{noise}(\rho)=\sqrt{\rho.(1- \rho)^{-1}}$. The noise distribution is free of learnable parameter. 
The categorical cross-entropy loss function $\mathcal{L}_{ce}(Y_{true}, Y_{pred})$  minimizes the error rates between the actual class-label ($Y_{true}$) and predicted class-label ($Y_{pred}$) during the learning task. The proposed SYD-Net is end-to-end trainable and the attention module could be added with standard backbones to  enhance efficiency. 

\subsection{Attention-based Baseline Method} 
{In addition to the conventional baseline evaluation, the attention mechanism is applied to compute  baseline results (Table \ref{baseline}). We aim  to observe the suitability of hybrid patches in improving the accuracy of attention-based baseline results using different backbone CNNs. A pictorial representation of the attention-based baseline is shown in Fig. \ref{fig:Attn_Baseline}. In this method, firstly, the high-level feature map from a backbone CNN is extracted. Subsequently, the self-attention and weighted attention techniques are exploited for re-weighting the base CNN's output features. Lastly, a softmax layer is applied  to the attentional feature description for classification.}

In summary, three methods are explored for baseline assessment: (a)  simple classification method using high-level feature vector of a base CNN with conventional data augmentation; (b) similar classification strategy as (a), with additional random erasing data augmentation for more data-diversity; and (c) leveraging attentional weights over base CNN's features in conjunction with (b), shown in Fig. \ref{fig:Attn_Baseline}.  Description of baseline evaluations is given in Sec. \ref{PerAna}. Indeed, our attention-based baseline performance  outperforms the traditional baseline method that only uses  base CNN's output features.
\vspace{-0.3 cm}
\section{Dataset Description} \label{datades}
We have described the Sports-102, Yoga-82, Yoga-107, and Dance-12 datasets which provide only class labels, avoiding bounding-box annotations, and summarized in Table \ref{DB}.

\subsubsection{Sports-102 Dataset}
The sports dataset represents 102 sports-action classes representing  complex human body postures. Sports-102  comprises various  games, some of which are based on individual performers (\textit{e.g.}, golf, javelin, etc.) while others are team-based (\textit{e.g.}, hockey, kabaddi, etc.) with diversity. Samples of various sports are shown in Fig. \ref{fig:s102}. The training and testing data distribution of various sports categories are shown in Fig. \ref{fig:data_Sports}. Mainly, the images are collected from Kaggle {\footnote{{www.kaggle.com/datasets/gpiosenka/sports-classification}}}  repository, and related websites. Though  few video-based datasets exist for dance and sport actions, no such image-based datasets are publicly available for research, to the best of our knowledge. 

\subsubsection{Yoga-82 \cite{verma2020yoga} and {Yoga-107 Datasets}}
These are publicly  available datasets. Samples of various  postures of Yoga-82 are illustrated in Fig. \ref{fig:yoga82_samples}.a. After careful observation, a few samples are rejected, which are irrelevant. The reason might be the resolution, format, size, other characteristics of images, and repetition of the same images. Some poses ($asana$) representing the cartoon's and animal's images  are irrelevant to the current problem, so, eliminated. Thus, samples from various yoga classes are discarded to formulate a well-defined and precise Yoga-82 sub-dataset. Actual Yoga-82 contains 21.0k training and 7.4k testing samples. Whereas we have tested on 19.9k training and 7.2k testing images after standardization. 

The Yoga-107 dataset is collected from Kaggle {\footnote{{https://www.kaggle.com/datasets/shrutisaxena/yoga-pose-image-classification-dataset}}} repository and a few samples are shown in Fig. \ref{fig:yoga82_samples}.b. It contains 107 fine-grained classes of yoga poses, comprising a total of 5.9k images. It is a challenging yoga dataset as the classes are more than 100,  and the $asana$ samples per class are much lesser than Yoga-82. 

\begin{figure*}[h]
\centering
\subfloat[Ballet]{ 
\includegraphics[width=0.12\linewidth, height= 1.8  cm]{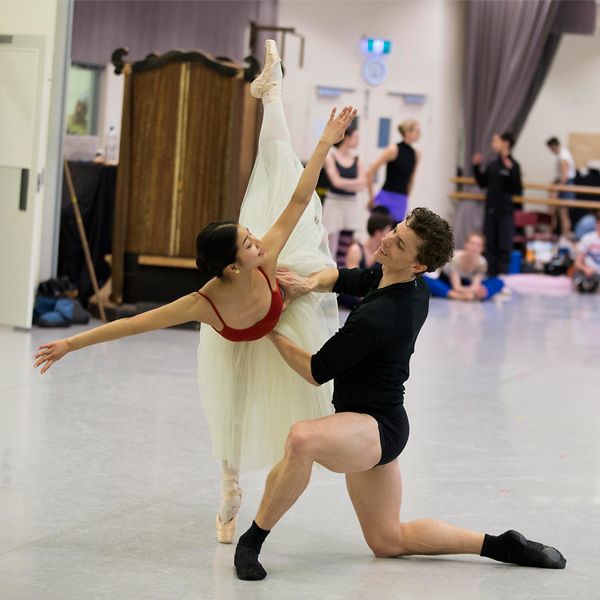} 

}
\subfloat[Hip-Hop]{ 
\includegraphics[width=0.11\linewidth, height= 1.8  cm]{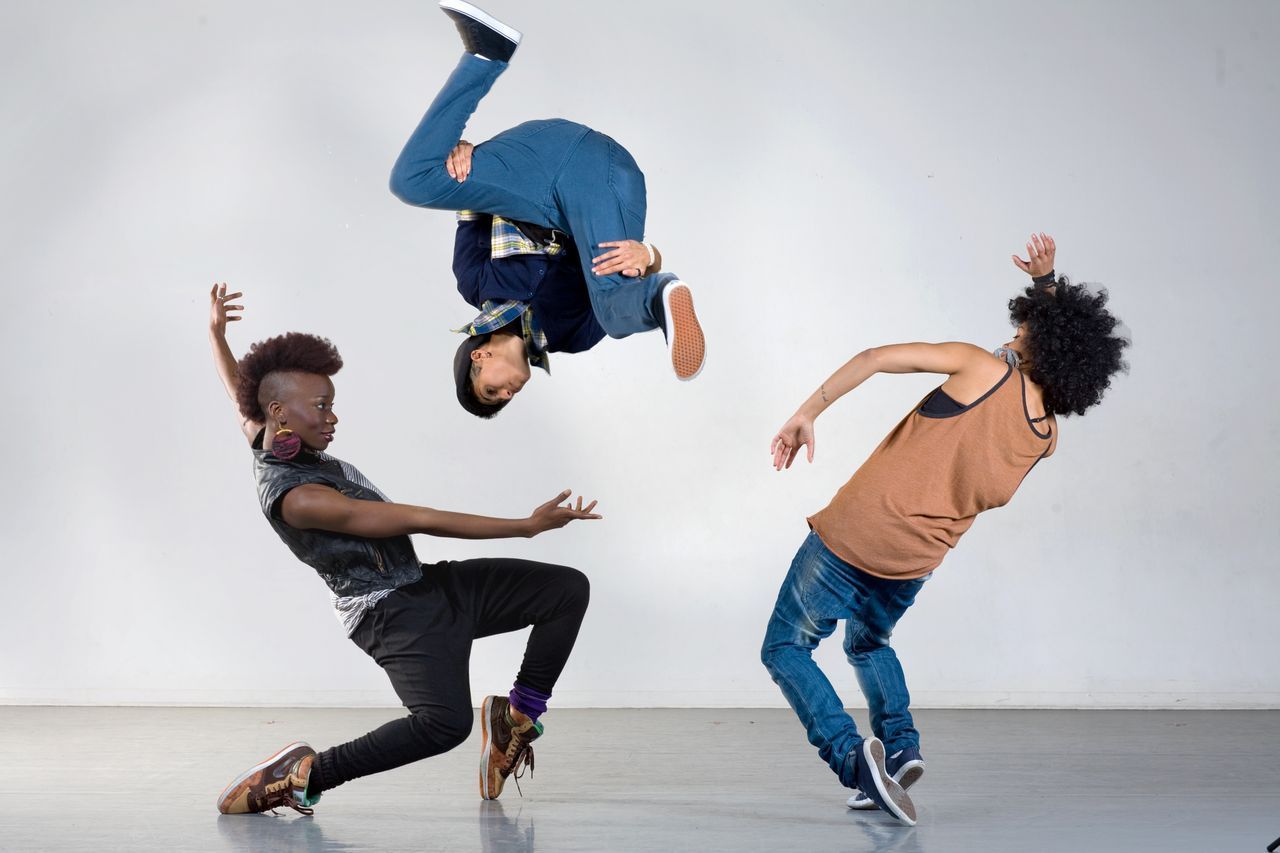} 

} 
\subfloat[Salsa]{ 
\includegraphics[width=0.11\linewidth, height= 1.8  cm]{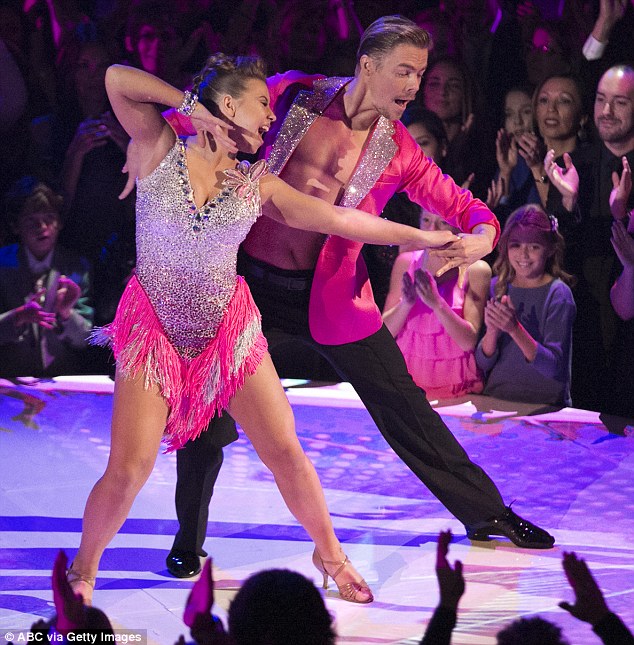}

} 
\subfloat[Pole]{ 
\includegraphics[width=0.11\linewidth, height= 1.8  cm]{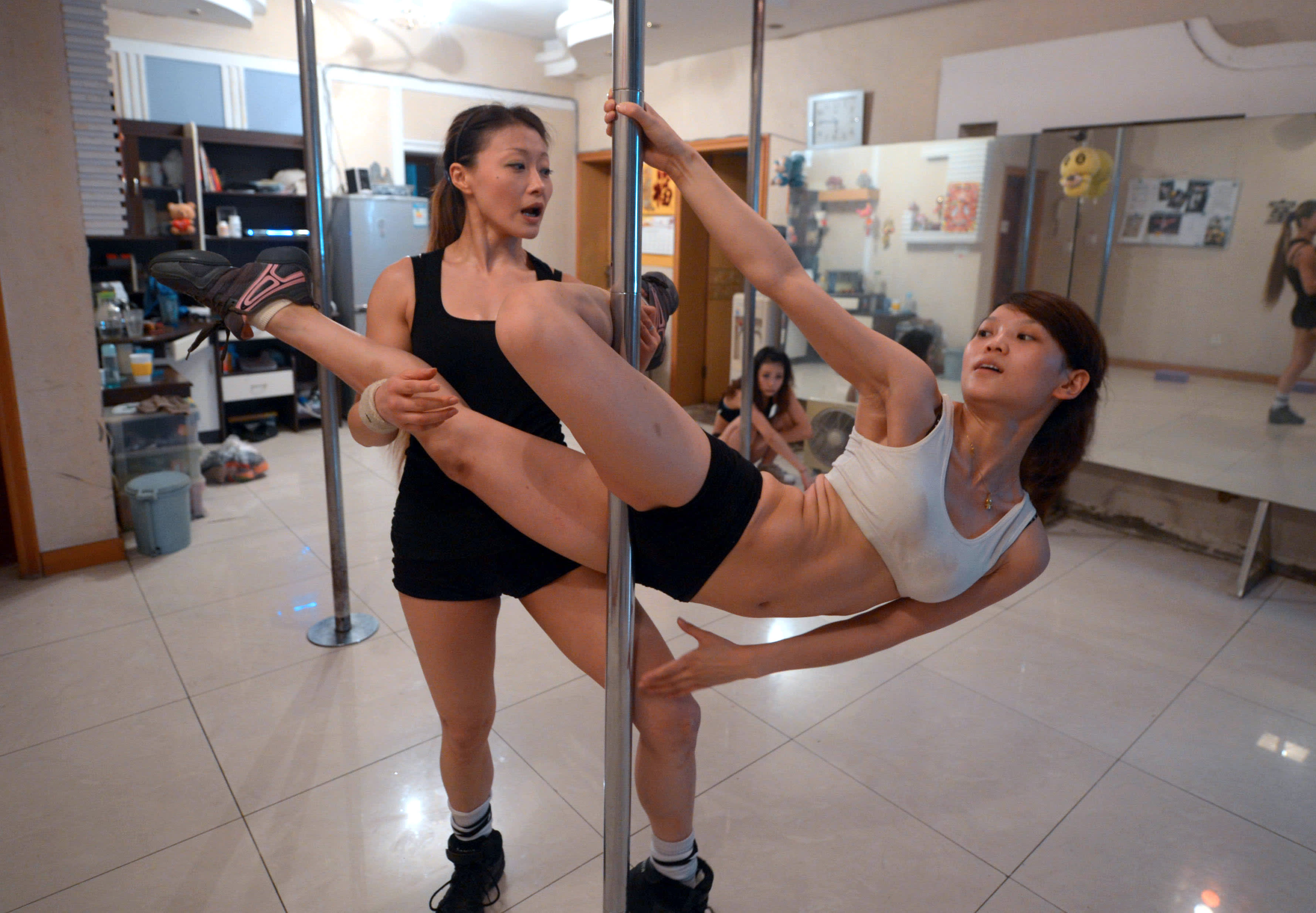}
} 
\subfloat[Bharatnatyam]{ 
\includegraphics[width=0.11\linewidth, height= 1.8  cm]{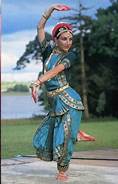} 
}
\subfloat[Manipuri]{ 
\includegraphics[width=0.11\linewidth, height=1.8  cm]{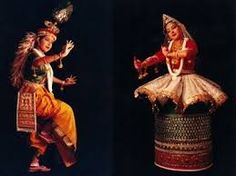} 
} 
\subfloat[Chhau]{ 
\includegraphics[width=0.12\linewidth, height=1.8  cm]{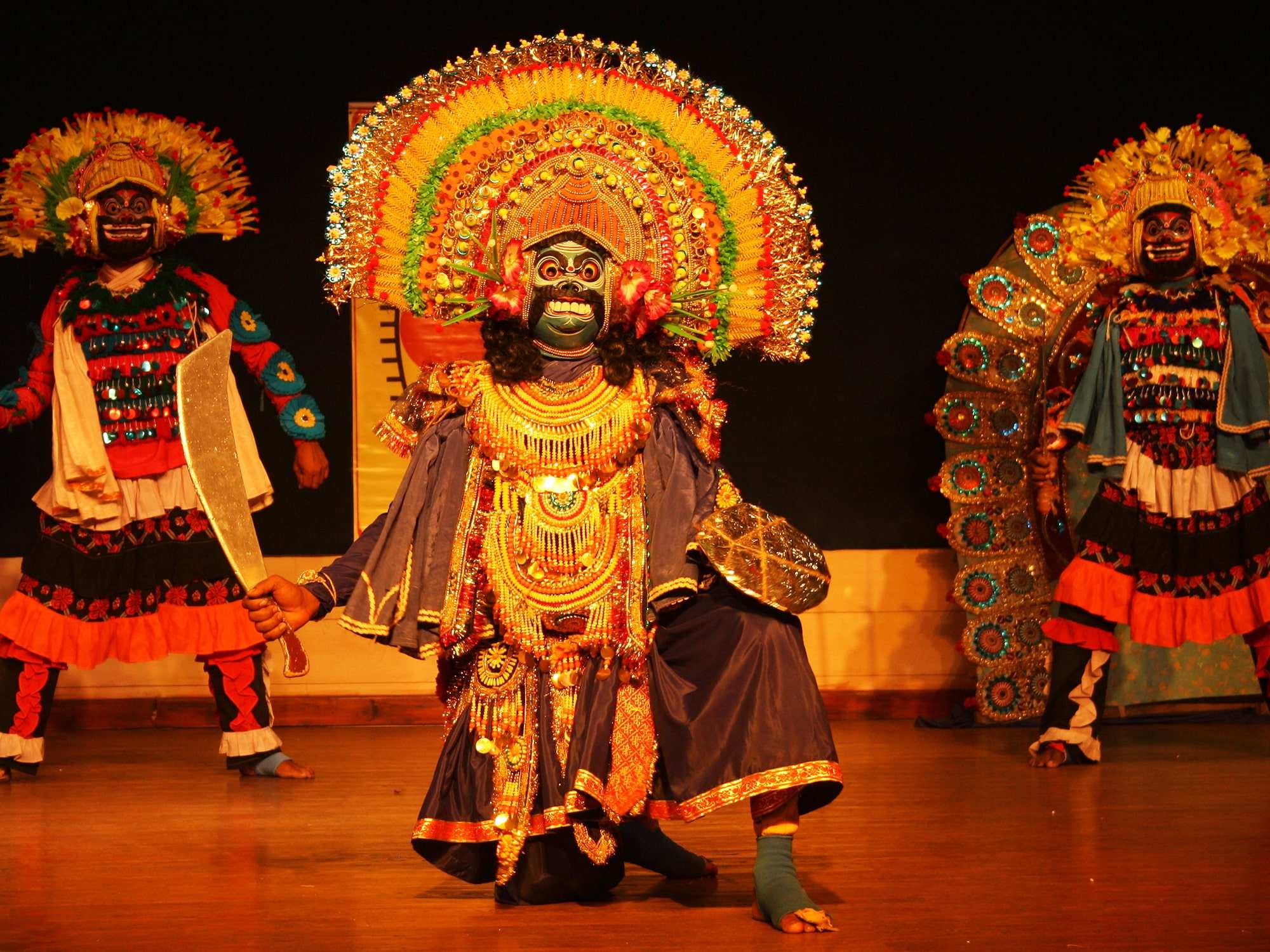}
} 
\subfloat[Kalbelia]{ 
\includegraphics[width=0.12\linewidth, height= 1.8  cm]{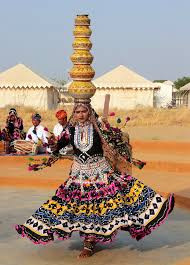}
} 
\vspace{-0.1 cm}
\caption{Examples of various dance styles from the Dance-12 dataset.}
\label{fig:DanceStyles}
\end{figure*}
\begin{figure}[h]
\centering
\includegraphics[width=0.7\linewidth, height= 3 cm]{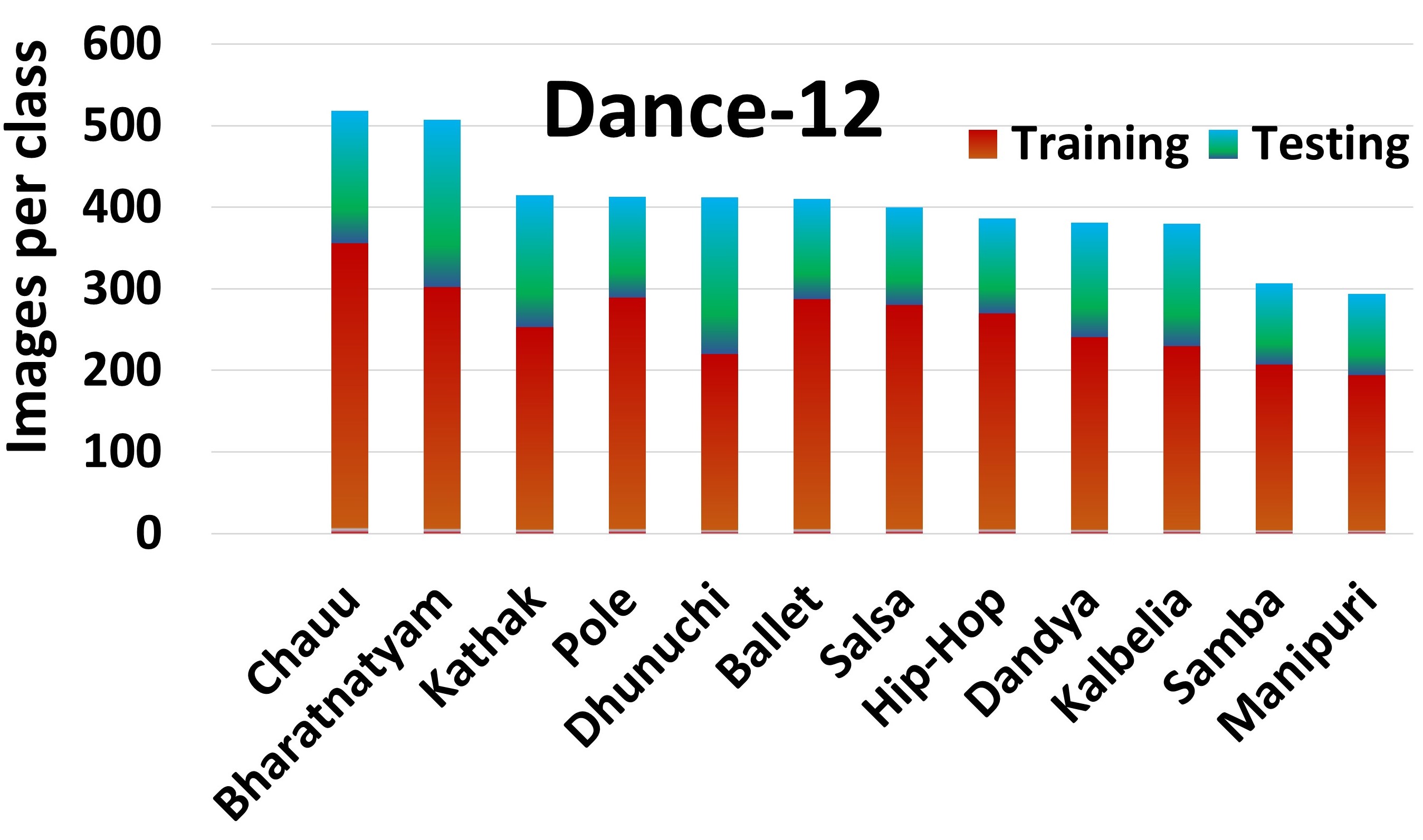} 
\caption{Training-testing image distribution of Dance-12 dataset.}
\label{fig:data_Dance}
\vspace{- 0.3 cm}
\end{figure}
\begin{table}
\vspace{-0.2cm}
\caption{Dataset Summary and Top-1 Accuracy (\%) of SYD-Net Trained from Scratch and with ImageNet Weights using Xception.}
\begin{center}
\begin{tabular}{|c|c c  c|c c|}
\hline
Dataset & Train & Test & Class & Xcep (srth)  & Xcep (ImNet)  \\
\hline
Sports-102  & 9278 & 4315 &102 & 96.70 & 98.86 \\
Yoga-82  & 19941 & 7241 & 82 & 97.29 & 97.80 \\
{Yoga-107}  & {4084} & {1883} & {107} & { {82.00}} & 85.20 \\
Dance-12  & 3129 & 1694 & 12 & 92.24 & 97.98  \\
\hline
\end{tabular}
\label{DB}
\end{center}
\vspace{-0.3 cm}
\end{table}

\subsubsection{Dance-12 Dataset}
A total of 12 dance styles are incorporated in the Dance-12  dataset representing diverse variations in postures, number of persons, background, theme, and other factors. This dataset represents the following dance categories: Ballet, Hip-hop, Pole, Salsa, Samba,  Bharatnatyam, Chaau, Dandya, Dhunuchi, Kathak, Kalbelia, and Manipuri. The first five dances are internationally popular and the remaining seven are Indian. All the images are collected freely from various public websites such as Google, Yahoo, Bing, etc. Samples of international and Indian dance genres are illustrated in Fig. \ref{fig:DanceStyles}. 
The training and testing splits of various dance styles are given in Fig. \ref{fig:data_Dance}. The purpose of our data collection is research only. No commercial gain or unethical issue is involved in our research. Dance-12 is growing a dataset in size and variations. We will include several more  classical and folk dance styles from various countries around the world in the near future. { This dataset is publicly available  at \texttt{https://sites.google.com/view/syd-net/home}}.

\section{Experimental Results} \label{expmnt}
Firstly, we have analyzed the experimental details of SYD-Net. Next, an ablation study is presented  to evaluate the significance of key components of the SYD-Net model. 
\subsection{Implementation} 
Our model is implemented using  ResNet-50 \cite{he2016deep},  DenseNet-201 \cite{huang2017densely}, NASNetMobile \cite{zoph2018learning}, MobileNet-v2 \cite{sandler2018mobilenetv2}, and Xception \cite{chollet2017xception} backbone CNNs in TensorFlow-2.x with cuDNN 7.6.{The input image resolution is 224$\times$224, and the output feature map of the MobileNet-v2 backbone is 7$\times$7$\times$1280. Whereas, the output feature maps of other backbones represent the same spatial size, but differ in channel dimensions. The patch sets are  extracted along the spatial dimension \cite{he2015spatial}. The spatial size of base output (7$\times$7) is upsampled to 48$\times$48 for extracting the patch sets from $P_{9}$ to $P_{20}$. The upscaled resolution is 45$\times$45 for proper pixel alignment with the patch-sizes of  $P_{25}$ and $P_{30}$.} Three sets of uniform  patches (3$\times$3, 4$\times$4, and 5$\times$5) and corresponding hierarchical regions are generated. For example, with 16 uniform patches, 4 multi-scale patches are computed in a hierarchical manner from the center of the input image with the smallest 12$\times$12 size, and incremented to  24$\times$24, 36$\times$36, and finally to 48$\times$48 size. Altogether 16 uniform and 4 hierarchical patches are considered in set $P_{20}$. Likewise, $P_{12}$ and $P_{30}$ are generated.
Initially, the input image size is 256$\times$256. We have applied standard data augmentations of random rotation ($\pm$25 degrees) and random scaling (1$\pm$0.25). Two randomly selected regions of total size or a single region with size (0.1-0.8) are erased with either a fixed RGB=127 or random RGB pixel-values at a time (Fig. \ref{fig:patches}), and applied on-the-fly for image augmentation. Then, random cropping is applied to select an image size of 224$\times$224 as input to CNNs. SYD-Net is trained from scratch for initializing  base CNNs for a fair comparison, as well as trained with ImageNet weights in separate experiments. The Stochastic Gradient Descent (SGD) is used to optimize the categorical cross-entropy loss function with an initial learning rate of $0.007$ and multiplied by 0.1 after every 50 epochs. The model is trained for 200 epochs with a mini-batch size of 8 using a Tesla M10 GPU of 8 GB. A Gaussian dropout rate 0.2 and batch normalization are applied to avoid overfitting. The top-1 and top-5 accuracy (\%) metrics are used for performance evaluation, and the model's parametric complexity is estimated in millions (M).

\vspace{-0.2 cm}
\begin{table}
\caption{Top-1 Baseline Accuracy (\%) using  Conventional  Data Augment (top row-set), Random  Erasing  Augment (mid row-set), and Attention with Random Erasing (last row-set). The last column shows Model Parameters in Millions (M).  
}
\begin{center}
\begin{tabular}{|c|c c c c|c|}
\hline
Base CNNs    & Sports & Yoga-82 & {Yoga-107} & Dance & Par (M) \\
\hline
ResNet-50 & 68.34  & 75.52 & {52.13}  & 63.44  & 23.8\\
DenseNet-201  & 74.21 & 80.16 & {55.12}   & 68.12 & 18.5  \\
MobileNet-v2  & 75.70 & 79.73 & {60.31}   & 63.80 &2.4 \\
Xception &  {79.10}   & {81.93} & {62.87}   & 72.27   & 21.1 \\
\hline
ResNet-50 & 70.91  &77.56 & {55.28}   & 63.80 & 23.8 \\
DenseNet-201 & 76.69 & 80.60 & {57.47}   & 68.24 &18.5 \\
MobileNet-v2 & 77.62 & 82.23 &  {65.10}   & 65.58 &2.4 \\
Xception & {80.17}   & {83.78} & {66.50}   & 72.92  & 21.1 \\
\hline
ResNet-50 &  {70.96} & {80.60} & {57.31}   & {60.90}  &23.9 \\ 
DenseNet-201 & 77.18 & 83.63 & {58.38}   & 68.95 &18.6 \\
MobileNet-v2  & 77.22 & 83.31 & {65.86}   & {68.06}  & 2.4\\
Xception & \textbf{82.88} & \textbf{85.67} & \textbf{67.52}  & \textbf{73.34}  & 21.2 \\
\hline
\end{tabular}
\label{baseline}
\end{center}
\vspace{ -0.4 cm}
\end{table}

\subsection{Performance Analysis}  \label{PerAna}
The performances of SYD-Net have been evaluated considering several important aspects, discussed next.

\subsubsection{Baseline Results} First,  baseline performances of four backbone CNNs (trained from scratch) are computed on four datasets. The results are given in Table \ref{baseline}. Three sets of experiments are conducted for a baseline evaluation, as aforesaid. In the first set of experiments (top row set), conventional data augmentations,\textit{i.e.}, rotation, scaling, and cropping are applied. In addition to the general augmentation,  random region erasing is applied in the second set of experiments, given in the middle row set. In both experiments, we  considered the output feature maps of base CNNs, and then applied global average pooling (GAP) prior to classification layer. In the last set of experiments (bottom row set), the erasing-based data augmentations remain the same. 
 
 Moreover, the attention module is applied as an alternative to GAP on base CNNs feature maps. However, the patches are not included in any baseline assessment (Fig. \ref{fig:Attn_Baseline}). The baseline performances have been improved using random erasing  over traditional augmentation techniques. Also, attention has enhanced the baseline accuracy over GAP with a little overhead regarding the model parameters (approx +83K).  

\begin{table} 
\begin{center}
 \caption{Top-1 Accuracy (\%) of SYD-Net (Scratch) with Attention  Modules  using  Various Uniform (U) and Hierarchical (H) Patches: $P_{9}$, $P_{12}$, $P_{16}$, $P_{20}$, $P_{25}$, and $P_{30}$. Two Random Erased Regions with Basic Image Augmentations are Applied.  }
 \label{table:RoI_Acc}
\begin{tabular}{|c c|c c c c|c| }
 \hline
 {CNNs} & Patch & {Sports} & Yoga82 & {Yoga107}  &  Dance  & Par(M)   \\ 
\hline
 RN-50  & $P_{9}$ & 88.10 & 91.43 & {70.19} & 78.61  & 32.0  \\ 
 &    $P_{12}$  & {89.37} & {92.56} & {70.67}  & {79.03} &  32.7 \\
 &   $P_{16}$  & 89.51  & 93.49 & {71.03} & {81.33} & 33.4 \\
 &    $P_{20}$  & 90.23 & 93.77 & {71.90} & {82.52} & 34.2  \\ 
   &    $P_{25}$  & 90.51  & {93.56} & {73.02} & {82.40} & 35.2 \\
   &    $P_{30}$  & 91.37  & {93.92} & {75.00}  & {84.18} & 36.2 \\

\hline
DN-201  & $P_{9}$ & {90.60} & {94.24}& {68.32} & 83.42 & 26.1 \\ 
 &  $P_{12}$  & {91.32} & {94.94} & {69.71}  & {83.00} &26.8  \\
 &    $P_{16}$   & 92.55 & 95.48 & {71.90} & {83.94} & 27.6 \\
 &   $P_{20}$   & 92.85 &  95.75 & {72.32} & {85.24} & 28.3  \\
    &  $P_{25}$  & {92.83} & {95.46} & {73.50} & {85.37} & 29.2 \\
   &    $P_{30}$  & {93.36} & {95.87} & {74.73} &  88.03 & 30.0 \\
\hline
MN-v2  & $P_{9}$ & {93.50} & 94.66 & {75.26} & 84.36 & 7.5 \\ 
 &    $P_{12}$  & 93.20 & 95.02 & {76.22}  & {84.60} & 7.9 \\
 &    $P_{16}$   & 92.90 & 95.06 &{77.67}  & {85.42} & 8.4 \\
 &    $P_{20}$  & 93.92  & 95.63 &{78.10} & 86.20 & 8.9 \\
  &   $P_{25}$  & 93.22 & 94.96 & {78.84} & 86.85 & 9.6 \\
   &  $P_{30}$  & 94.78 & {96.00} & {79.54} & 87.73 & 10.2 \\
\hline
XN  & $P_{9}$ & {95.66} &  96.35 & {71.36} & 87.50 & 29.3 \\ 
 &  $P_{12}$  & {95.89} & {96.68} & {77.72} & {89.21} &{29.9}   \\
 &   $P_{16}$  & {95.40} & {96.40} &{79.54} & {90.10} & {30.7} \\
 &   $P_{20}$   & 96.03 &  96.76 & {80.19} & 90.22 & 31.5  \\
    &  $P_{25}$  & {96.21} & 96.50 & {80.76} & 91.40  & 32.5 \\
   &    $P_{30}$  & \textbf{96.70} & \textbf{97.29} & {\textbf{82.00}} & \textbf{92.24} & 33.5 \\
\hline
\end{tabular}
\end{center}
\vspace{ -0.4 cm}
\end{table}

 \begin{table}
{
\caption{SYD-Net's Top-1 Accuracy(\%) using NASNetMobile (Scratch)}
\begin{center}
\begin{tabular}{|c c c c c|c|}
\hline
Method &  Sports  &  Yoga-82   &Yoga-107  & Dance  & Param (M) \\
\hline
 Erasing BL & 70.82 & 77.61  & {56.20} & 66.35 & 4.4 \\
$P_{20}$  & 90.62 & 94.08  & {72.70} & 
  83.88 & 9.8 \\
$P_{30}$  & {91.58} & 94.70 & {74.09} &   84.12 & 12.4 \\  
\hline
\end{tabular}
\label{Y107NasMob}
\end{center}
}
\vspace{ -0.6 cm}
\end{table} 
\begin{table}[h]
\caption{Top-5 Accuracy(\%) of SYD-Net with  $P_{30}$, Trained from Scratch 
}
\begin{center}
\begin{tabular}{|c c c  c  c|}
\hline
CNNs &  Sports-102  & Yoga-82  & {Yoga-107}  &  Dance-12     \\
\hline
 ResNet-50 & 98.89 & 99.42 & {95.67}  &  98.22   \\
 DenseNet-201 & 99.40 & 99.20 & {96.20}  & 98.93\\
 NASNetMobile & {99.07} &  {99.60} & {95.99}  & {98.69 } \\
 MobileNet-v2  & 99.60  & 99.78 &{97.95} & 98.70 \\
 Xception &  99.81 & 99.42 & {98.39} & 99.70 \\ 
\hline
\end{tabular}
\label{top-5}
\end{center}
\vspace{ -0.3 cm}
\end{table}
 \subsubsection{SYD-Net's Performance}  
 The accuracy of SYD-Net is improved significantly by incorporating an attention module (PbA) over hybrid patches. The performances of a different number of patches using four base CNNs \textit{i.e.}, ResNet-50 (RN-50), DenseNet-201 (DN-201), MobileNet-v2 (MN-v2), and  Xception (XN) are evaluated on all four datasets, and the results are given in Table \ref{table:RoI_Acc}. It evinces that  uniform (U) patches could attain good results over baseline accuracy. Moreover, uniform patches in conjunction with hierarchical (H) regions boost the performance further by summarizing contextual descriptions at multiple granularities. We have defined three sets of mixed patches: $P_9$ contains 3$\times$3=9 (patch-size 16$\times$16 pixels),  $P_{16}$ represents 4$\times$4=16 (patch-size 12$\times$12 pixels), and  $P_{25}$ represents 5$\times$5=25 (patch-size: 9$\times$9 pixels) uniform (U) regions. It is clear that more patches attain better results. For example, $P_{25}$ renders better results than $P_{9}$ and  $P_{16}$. To enhance the accuracy of uniform patches  further, 3 hierarchical regions are included with $P_{9}$  to produce $P_{12}$ (9U+3H) hybrid patches. Likewise, 4 multi-scale regions are included with $P_{16}$ to generate $P_{20}$ (16U+4H) patches, and 5 multi-scale regions are included with $P_{25}$ to generate $P_{30}$ (25U+5H) patches, respectively. The results imply that  hybrid regions could improve accuracy over  all three  ($P_{9}$, $P_{16}$, and $P_{25}$) sets of uniform patches. Finally, set $P_{30}$ achieves the best performance among all patch sets.
MobileNet-v2 achieves competitive results compared to heavier backbones. Thus,  another light-weight CNN, NASNetMobile, is tested  on these datasets additionally.  Only the baseline with random erasing augmentation (Erasing BL), $P_{20}$, and $P_{30}$ are considered in this precise experiment. The results  of NASNetMobile trained from scratch are given in Table \ref{Y107NasMob}. Both  MobileNet-v2 and NASNetMobile, albeit lightweight,  have attained competitive accuracy over other base CNNs. 

Next, the top-5 accuracy (\%) of SYD-Net using $P_{30}$ with two random erased regions trained from scratch are given in Table \ref{top-5}.  All backbone CNNs attain excellent top-5 accuracy on four SYD datasets.
Now, SYD-Net is trained  with  pre-trained \textit{ImageNet} weights to observe its efficiency. This experiment is conducted with both lightweight CNNs, and the best performer Xception considering random erasing augmentation. The results are given in Table \ref{TWORandEraseRoI}. The accuracy is improved by a significant margin using \textit{ImageNet} weights on  $P_{20}$ and $P_{30}$ compared to training from scratch (Table \ref{table:RoI_Acc}). Also, $P_{30}$ offers a little accuracy gain over $P_{20}$. 

Furthermore, SYD-Net is tested on the Yoga-107 dataset trained with \textit{ImageNet} weight initialization using four base CNNs, considering all combinations of patches and random erasing data augmentation, as defined above.  The top-1 accuracies are provided in Table \ref{Yoga107}. The best performance on Yoga-107  is 87.17\%, achieved by DenseNet-201 with $P_{30}$ patches. Also, other CNNs have achieved competitive results, \textit{e.g.}, Xception achieved the second-best accuracy of 85.20\%.

\begin{table}[h]
\caption{Top-1 Accuracy (\%) of SYD-Net trained with ImageNet weights using two erased regions with random values}
\begin{center}
\begin{tabular}{|c|c|c c c c|}
\hline
CNNs & Model   & Sports & Yoga-82 & {Yoga-107}    & Dance  \\
\hline
{NASNetMobile}  &  {BL} & {91.53}   & {88.48} &{73.34}  & {88.27}   \\
& {$P_{20}$}  & {97.56}  & {94.44} & {83.92}  & {95.80} \\
 & {$P_{30}$} & {98.21}  & {96.58} & {84.61}  & {96.74}\\
\hline
MobileNet-v2  &  BL & 93.11 & 90.73  & {75.64}   & {87.38} \\
 & $P_{20}$  &  {98.37} & 97.20  & {83.60} & {96.50} \\
  & $P_{30}$ & {98.70}  & 97.41 & {85.14}  & {96.80}  \\
\hline
Xception  &  BL &  {94.48}  & 91.29 & {73.34}  &  {88.74} \\
& $P_{20}$  &  {98.81}  & 97.41 & {84.18}  & {97.15} \\
 & $P_{30}$ &  {98.86}  & {97.80} & {85.20}  &{97.98} \\
\hline
\end{tabular}
\vspace{-0.5 cm}
\label{TWORandEraseRoI}
\end{center}
\end{table}
\begin{table}[h]
\vspace{-0.2 cm}
\caption{Top-1 Accuracy (\%) of SYD-Net Trained with {ImageNet} Weights on Yoga-107. The Terminologies are Denoted as: RN50 for ResNet-50; DN201 for DenseNet-201; XN for Xception; MNv2 for MobileNet-v2;  BL for Baseline; and $P_i$ for Patches. }
\begin{center}
\begin{tabular}{|c|c|c|c|c|c|c|c|}
\hline
CNN & BL   & $P_{9}$ & $P_{12}$ & $P_{16}$   & $P_{20}$   & $P_{25}$    & $P_{30}$ \\
\hline
RN50 & 74.20 &  80.76  &81.99  &82.42   & 83.11 & 83.60 &85.04  \\
DN201 & 77.83 &  83.06  & 85.84  & 86.48  & 86.54  & 86.75 & \textbf{87.17}  \\
MNv2  & 75.64  & 79.22 & 80.87  &82.66 & {83.60} & 84.56 & 85.14\\
XN  & 74.41 &  82.21  & 83.65  & 83.97  & 84.18 & 84.33 & {85.20}  \\
\hline
\end{tabular}
\vspace{-0.3 cm}
\label{Yoga107}
\end{center}
\end{table}

\subsubsection{Model Parameters}   
The complexity of SYD-Net  in terms of model parameters estimated in millions (M) is given in the last column of  Table  \ref{baseline}-\ref{Y107NasMob}. {MobileNet-v2 and NASNetMobile  are lightweight backbones than other CNNs, regarding the model complexity.  The model parameters of $P_{9}$ and $P_{30}$ patch sets using MobileNet-v2  are 7.5M and  10.2M, respectively. Next, NASNetMobile's baseline parameters are 4.4M, and $P_{30}$ requires 12.5M. Thus, the parametric complexity of $P_{30}$ using NASNetMobile backbone is the second lowest (after 10.2M parameters of MobileNet-v2) among all five CNNs used here.} MobileNet-v2 attains balanced performances on SYD datasets with lesser model parametric overhead, whereas SYD-Net using Xception backbone has attained the best accuracy with higher model parameters (33.5M). The accuracies of other base CNNs are satisfactory, and useful for benchmarking the SYD datasets. 
\vspace{ -0.2 cm}
\begin{figure}[h]
\centering
\subfloat[Top-1 Accuracy on Yoga-107]{ 
\includegraphics[width=0.48\linewidth, height= 3.5 cm]{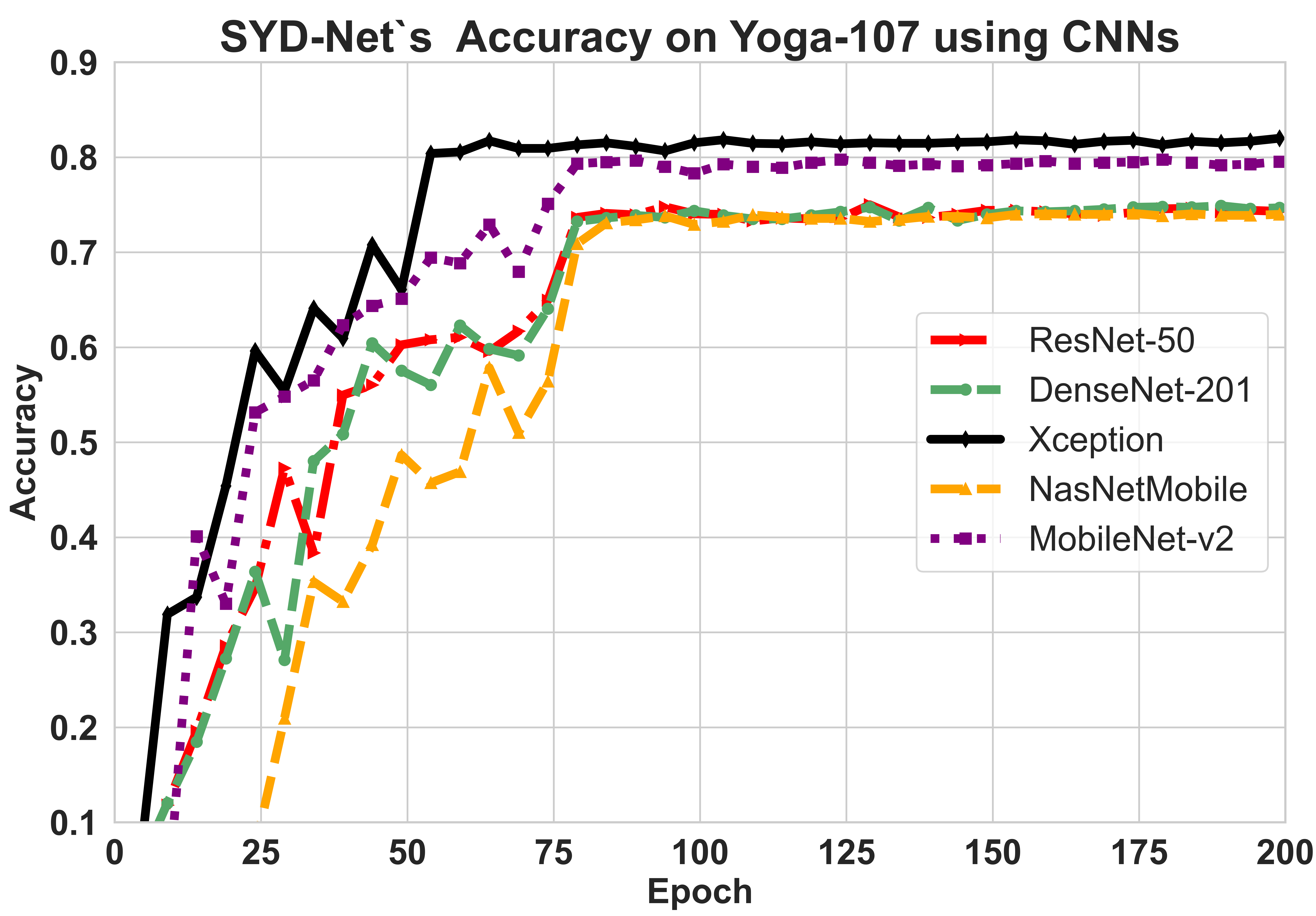} 
}
\subfloat[Accuracy Density]{
\includegraphics[width=0.48\linewidth, height= 3.5 cm]{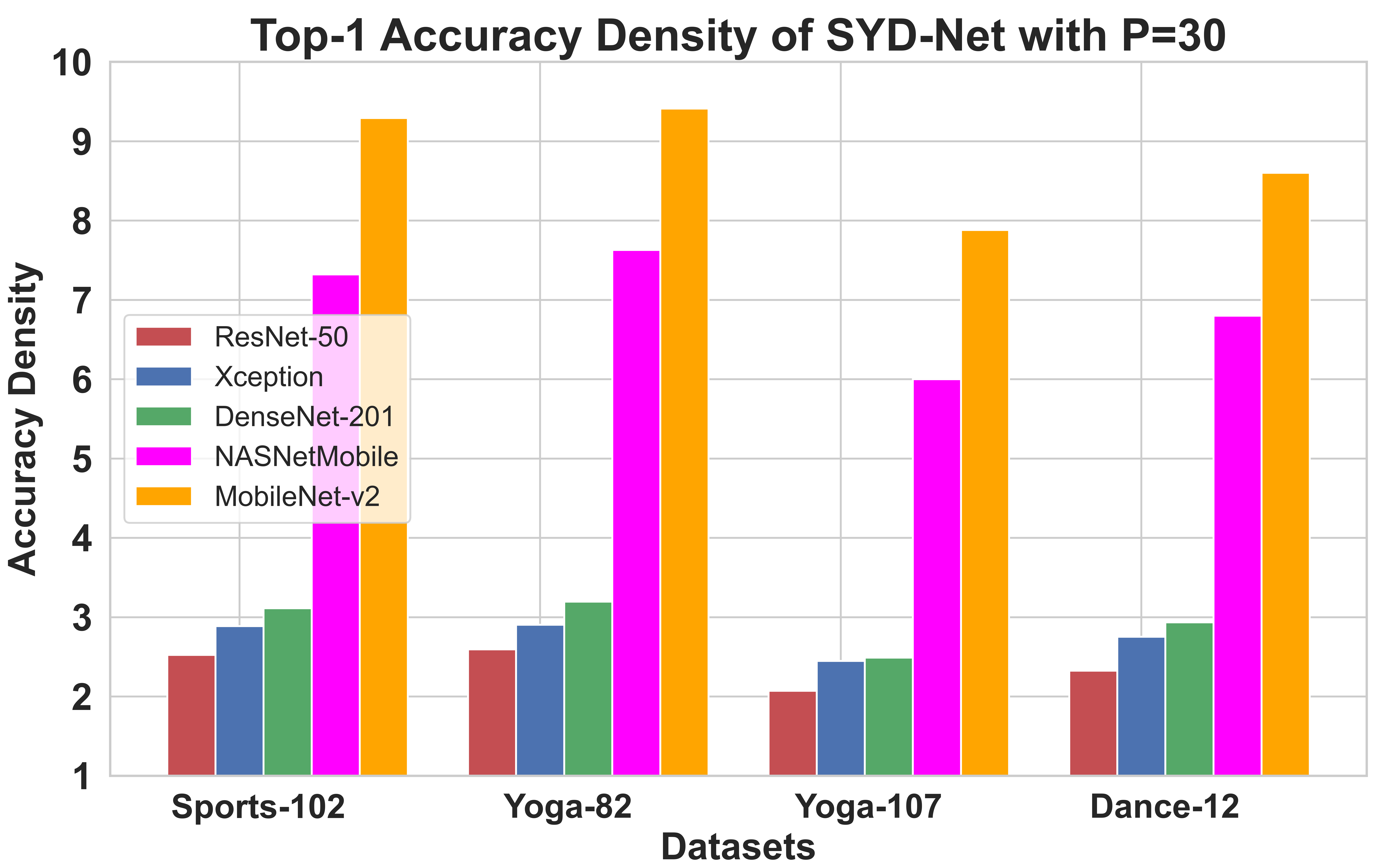} 
}
\caption{{Performance comparison of SYD-Net with $P_{30}$ using five different backbone CNNs trained from scratch. (a) Top-1 accuracy on Yoga-107. (b) Accuracy density on  four datasets using five backbone CNNs.} }
\label{fig:CompY107}
\vspace{-0.2 cm}
\end{figure}

\subsection{Comparative Study on Efficiency of Backbone CNNs}
Few works tested on small-scale datasets have witnessed  that MobileNet-v2 could achieve better performance than ResNet-50. A reason could be the fundamental design aspects of these standard backbones. ResNet family exploits shortcut connections (\textit{i.e.}, identity mapping) to avoid performance degradation problem. This conjecture is further improved by introducing bottleneck layers where parameter-free identity shortcuts are crucial in the network at a deeper level. On the contrary, Xception is hypothesized by utilizing the inception module and separable convolutions for decoupling the spatial and channel-wise feature correlations. MobileNet-v2 is built upon the depth-wise and point-wise separable convolutions and the inverted residual with linear bottleneck layers. Moreover, ReLU6 non-linearity is used for handling robustness issues at a lower dimensional feature representation. Overall, this  lightweight architecture directs to a faster and memory-efficient implementation than standard convolution, which is used as a main building block of other backbones. The compact, lightweight architecture of MobileNet-v2 leads to a higher computational  and accuracy gain over other backbone CNNs. A comparative study on Yoga-107 using five base CNNs is shown in Fig. \ref{fig:CompY107}.a, and Xception renders the best accuracy among all CNNs. The results indicate superior performances of both lightweight CNNs compared to other heavier backbones, ResNet-50, and DenseNet-201 \textit{i.e.}, densely connected between layers \cite{huang2017densely}. The results on Yoga-82, as reported in \cite{verma2020yoga}, imply that the MobileNet family outperforms the ResNet family. NASNetMobile attained competitive results \textit{i.e.}, the differences between the accuracies of NASNetMobile and ResNet-50 on various fine-grained datasets are small in \cite{bera2022sr}. Thus, the results reported in various works evince the better capacity of lightweight MobileNet-v2 and NASNetMobile base CNNs. Herein, the efficacy of both lightweight CNNs is clear from the baselines and patched-based results. 

To delve insight into the model’s capacity,  the performances of various base CNNs could be analyzed with the top-1 accuracy density, \textit{i.e.}, the ratio of top-1 accuracy and the number of model parameters, as defined in \cite{bianco2018benchmark}. A higher accuracy density value implies a higher efficiency of a deep network. It indicates  how efficiently the parameters contribute to the model’s capacity and expressiveness through successive layers of transformation and non-linearity. In \cite{bianco2018benchmark}, MobileNet-v2 attained better accuracy density than other base CNNs, \textit{e.g.}, ResNet-50. Our analysis of accuracy density follows a similar trend. A comparative study of accuracy density rendered by five base CNNs on four datasets is shown in Fig. \ref{fig:CompY107}.b. As aforesaid, a worthy reason of attaining a higher accuracy density is the powerful and precise architectural design of MobileNet-v2.
\begin{figure}[h]
\centering
\subfloat{ 
\includegraphics[width=0.47\linewidth, height= 4 cm]{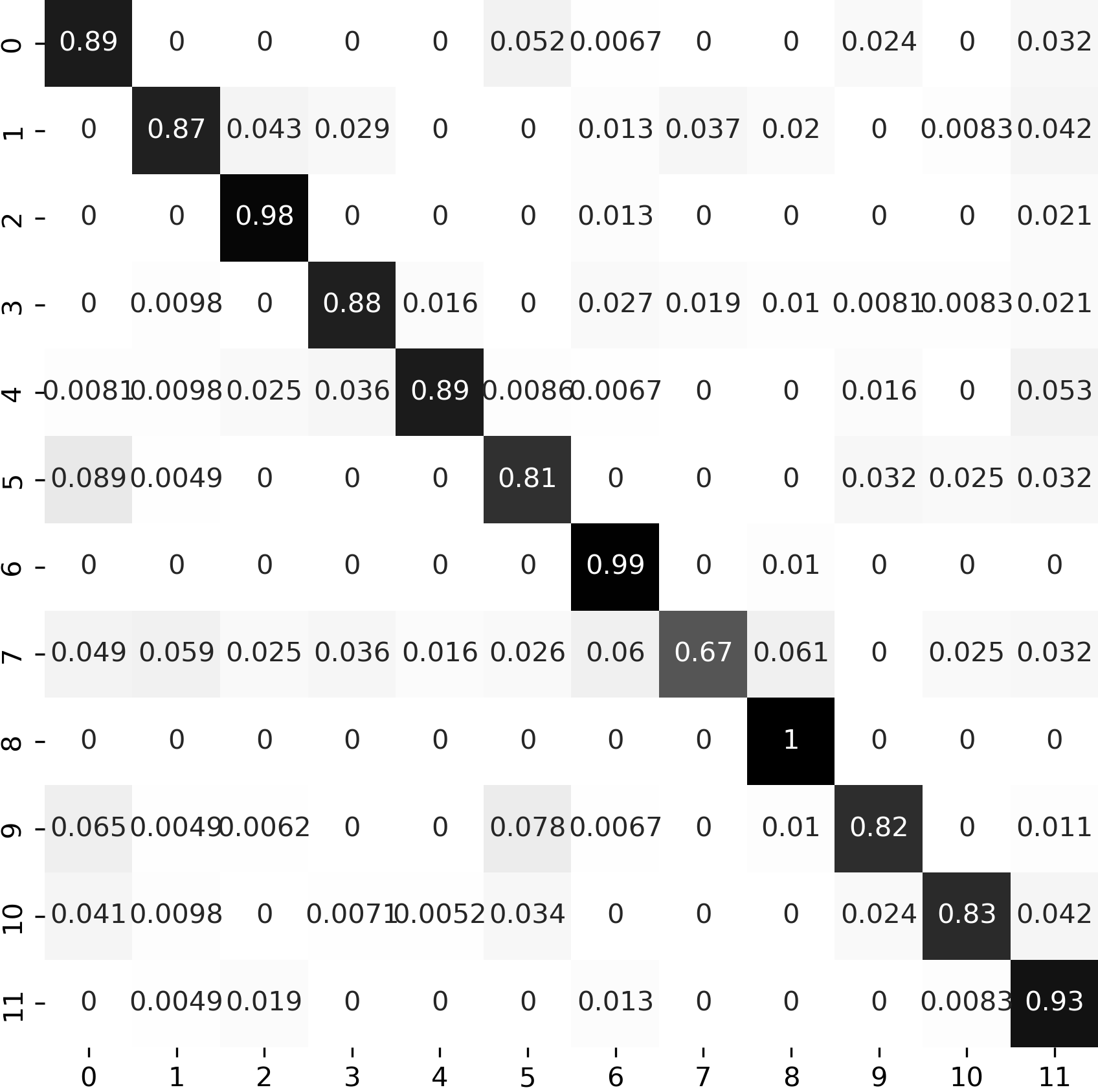} \hspace{0.2 cm}
\includegraphics[width=0.47\linewidth, height= 4 cm]{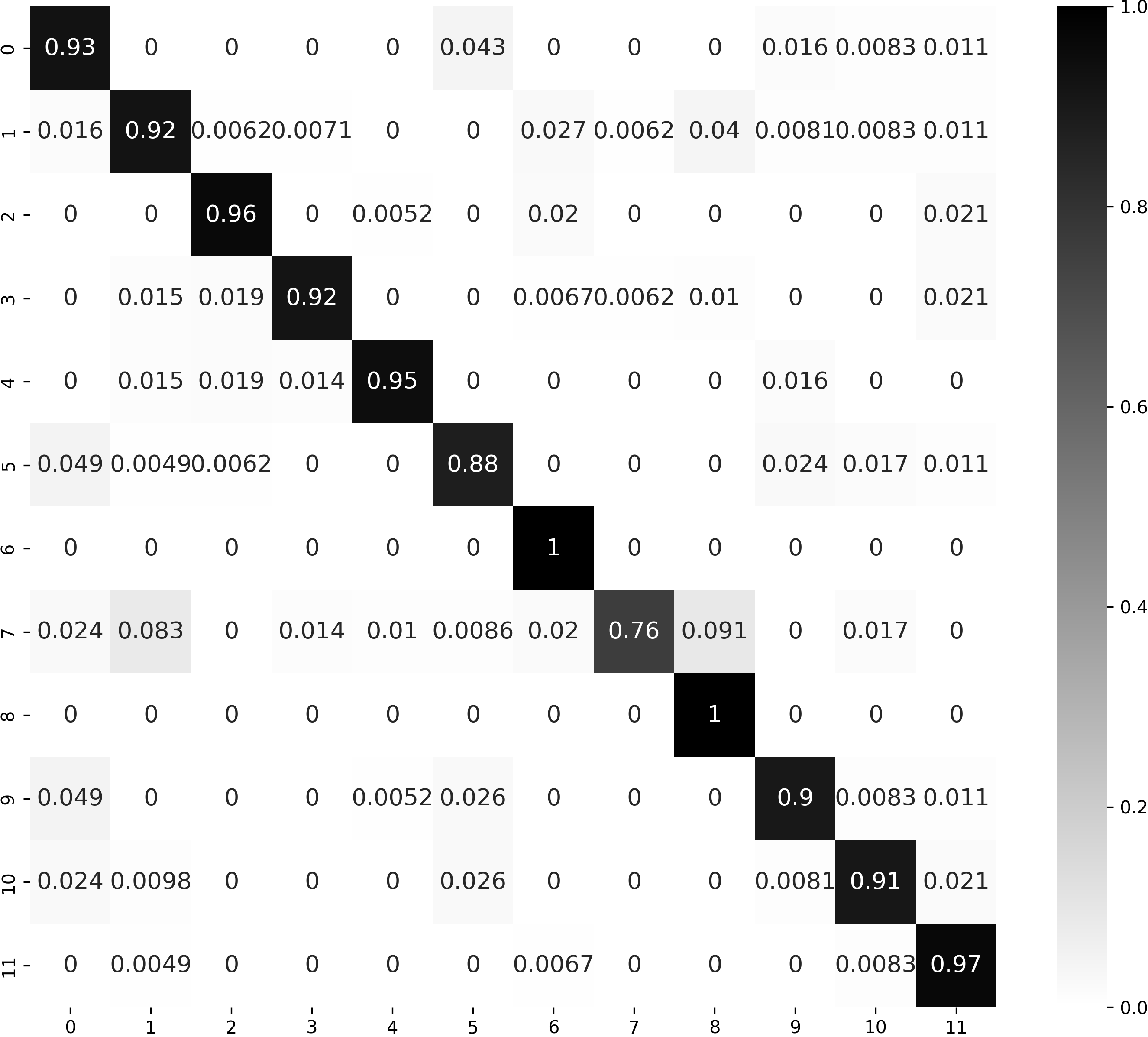} }
\caption{Confusion matrices of SYD-Net with $P_{30}$ on Dance-12 using MobileNet-v2, and Xception base CNNs. Best viewed in zoom.}
\label{fig:ConMat}
\vspace{ -0.5 cm}
\end{figure}
\begin{figure*}[h]
\centering
\subfloat{ 
\includegraphics[width=0.22\linewidth, height= 3. cm]{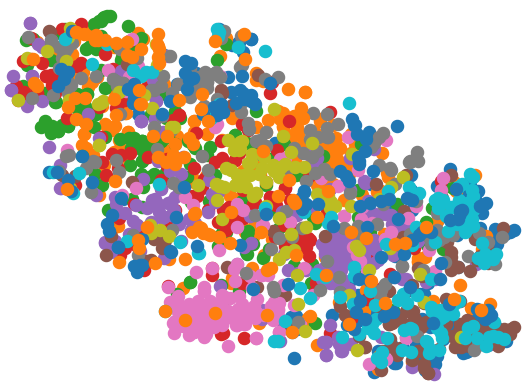} \hspace{0.3cm}
\includegraphics[width=0.22\linewidth, height= 3. cm]{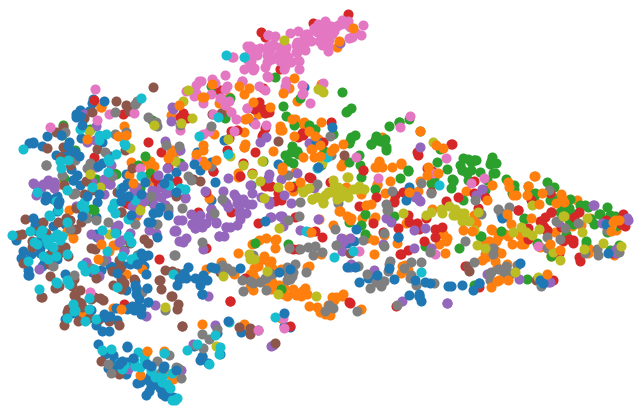} \hspace{0.3cm}
\includegraphics[width=0.22\linewidth, height= 3. cm]{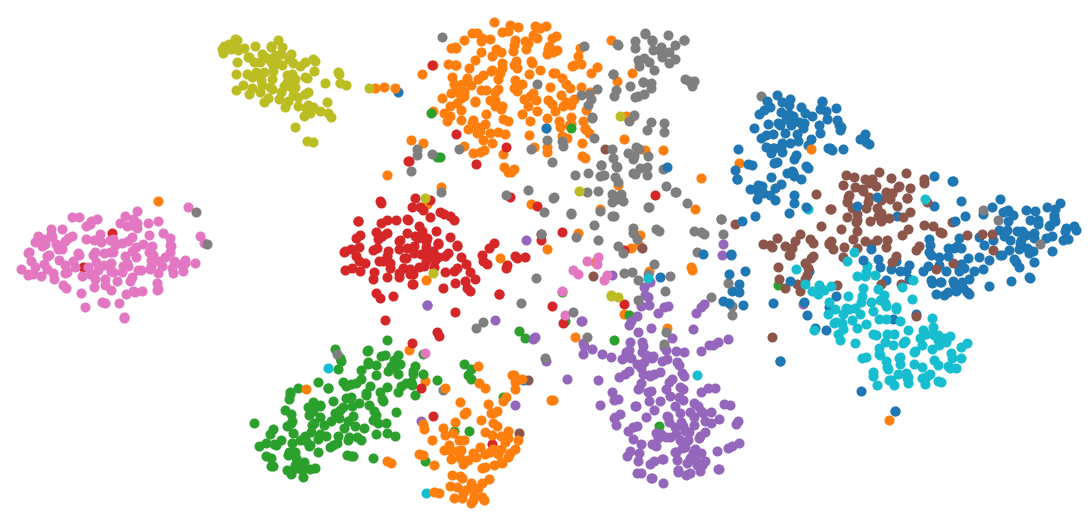} \hspace{0.4cm}
\includegraphics[width=0.22\linewidth, height= 3.cm]{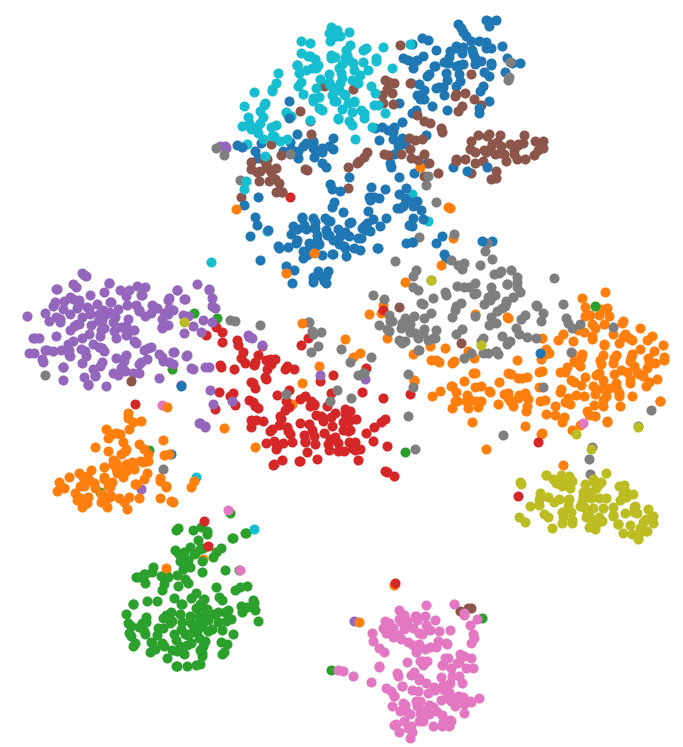}
}
\caption{The tSNE plots of SYD-Net on Dance-12 using MobileNet-v2. Left to right: General baseline; Baseline with attention mechanism; SYD-Net with $P_{25}$, and finally, SYD-Net with $P_{30}$. }
\label{fig:tSNE}
\vspace{ -0.4 cm}
\end{figure*}

\subsection{Performance Comparison on Yoga-82 and Yoga-107}
The pioneering work on  Yoga-82, considering 82 classes and using a variant of DenseNet-201,  has attained the best 79.35\%  top-1 and 93.47\%  top-5 accuracy, respectively. 
The performances on Yoga-82 using various CNNs are given in Table \ref{Comparson_Y82}. The best performances of our approach are compared  fairly using different CNNs trained from scratch. It is evident that SYD-Net outperforms those existing methods by a significant margin and achieves state-of-the-art results. The accuracies  of our baseline methods are higher than actual Yoga-82. As a result, the top-1 and top-5 accuracy of SYD-Net using various CNNs are significantly higher than existing works on Yoga-82. The body key-points-based classifier ensemble method has achieved 80.14\% top-1 accuracy  \cite{chasmai2022view}. Fusion of  DenseNet-161 and KNN model achieves 79\% accuracy while DenseNet-161 alone can attain 81\% accuracy in Pose tutor \cite{dittakavi2022pose}. On the contrary, SYD-Net has achieved at least a 14\% gain in top-1 accuracy using various CNNs.  

In Yoga-45 \cite{wu2022computer}, 1931 images representing 45 yoga classes were selected and achieved 83.27\% accuracy for pose grading using contrastive skeleton feature representation. In contrast, we have selected  5k images categorized into 107 classes as an enhanced version of Yoga-45. The best accuracy of SYD-Net is 87.17\%, rendered by DenseNet-201 with ImageNet weight initialization, reported in Table \ref{Yoga107}. However, our method is not directly comparable with Yoga-45 due to the differences in  dataset characteristics (\textit{e.g.}, sample size and the number of classes) and experimental setup.

The best top-1 accuracies of SYD-Net on four datasets using Xception, trained from scratch (srth), and ImageNet weights (ImNet) are reported in the last two columns of Table \ref{DB}. The proposed Sports-102 and Dance-12 datasets are publicly available  for further improvement and comparative analysis.

\begin{table}
 \vspace{-0.2cm}
\caption{Comparison of Top-1 and Top-5 Accuracy(\%)  on Yoga-82. }
\begin{center}
\begin{tabular}{|c|c|c|c|}
\hline
{Method} & {Backbone Types}  &  Top-1 Acc & Top-5 Acc  \\
\hline
Yoga-82 \cite{verma2020yoga}  & MobileNet-V2  & 71.11 &88.50 \\ 
& ResNet-50  & 63.44 &82.55 \\ 
& DenseNet-201 variant  & 79.35 & 93.47\\ \hline 
Ensemble \cite{chasmai2022view}   & keypoints + ensmble  & 80.14 & -\\ 
Fusion\cite{dittakavi2022pose}  & DenseNet-161 (ImNet) & 81.00 & - \\ \hline 
\textbf{SYD-Net}   & ResNet-50 & 93.92 & 99.42\\   
        & NASNetMobile & 94.70 & 99.60\\ 
   & DenseNet-201 & 95.87 & 99.20\\ 
   & MobileNet-v2  & 96.00 & \textbf{99.78} \\   
     & Xception & \textbf{97.29} & 99.42 \\ 
\hline
\end{tabular}
\vspace{-0.5 cm}
\label{Comparson_Y82}
\end{center}
\end{table}
\subsection{Feature Maps Visualization}
The confusion matrices of Dance-12  with $P_{30}$ are shown in Fig. \ref{fig:ConMat}. We have delved into various key layers to visualize feature maps using the t-SNE \cite{van2008visualizing} plots in Fig. \ref{fig:tSNE}. The figures show the feature distributions of data separability and clusters to reflect the discriminativeness of SYD-Net features. Here, the Dance-12 test set is considered for summarizing the feature distributions into a smaller subspace for visualization. In Fig. \ref{fig:tSNE}, the first two images show  a comparison of feature representation between the traditional data augmentation and its improvement using the attention mechanism. Both techniques are considered as baselines. The last two t-SNE figures show  feature representations of $P_{25}$ and $P_{30}$ in a lower dimension. These figures clearly show the class-wise feature map clusters with a significant class separability over the baselines. Also, the  data distribution  in  $P_{30}$ is slightly improved over $P_{25}$. This difference is reflected in the accuracy.

\subsection{Ablation Study}
The effectiveness of major components of SYD-Net is assessed on Dance-12 and Sports-102 using the MobileNet-v2 backbone. Mainly, the ablation study is focused on: (1) the patch-based attention module (PbA) and its sub-components; (2) the significance of two different activations in MLP; and (3) the impact of variations in random erasing data augmentations.  The results are reported in Table \ref{table:Abln2}.

\subsubsection{Patch-based attention module and its components}
We have evaluated the accuracy of three sets of hybrid patches without any attention module, \textit{i.e.}, $P_{12}$ (9U+3H), $P_{20}$ (16U+4H), and $P_{30}$ (25U+5H) regions. The results imply that the inclusion of patches could improve the accuracy over the baseline performances, given in Table \ref{baseline}. Also, $P_{30}$ renders better accuracy compared to $P_{12}$ and $P_{20}$. 

The significance of channel attention and spatial attention mechanisms of SYD-Net are explored. We have tested both paths separately to analyze the effectiveness of attention mechanism in a performance gain. It is evident that the patch-based channel attention (CA) path is more beneficial than the spatial attention (SA) module. The reason could be that feature maps optimization across the channel dimension (MobileNet-v2: 7$\times$7$\times$1280) is more effective over the spatial dimension (7$\times$7$\times$2). Because the feature space per patch is larger in cross-channel interaction than in spatial dimension, which ignores discriminative information during feature selection through spatial pooling. Also, $P_{30}$ achieves better accuracy than $P_{12}$, as observed in earlier ablation studies.

\begin{table} 
\begin{center}
 \caption{Accuracy (\%) of Various Key Components of SYD-Net and Random Erasing Augment using MobileNet-v2.}
 \label{table:Abln2}
 \vspace{-0.2 cm}
\begin{tabular}{|c |c  c|c| }
 \hline
 SYD-Net components  & Sports   &  Dance  & Par    \\ 
\hline
 Using  $P_{12}$ only, no attention   & {89.56}  & {69.78} & 2.4 \\ 
 Using   $P_{20}$ only, no attention   & {86.94}  & {68.54} & 2.4 \\ 
  Using   $P_{30}$ only, no attention   & {92.57}  & 80.21 & 2.4 \\ 
\hline
Spatial attention only: $SA_{12}$    & {77.18}  & 67.71 & 3.9  \\
 Spatial attention only: $SA_{20}$    & {77.36}  & 69.07 & 4.8  \\
 Spatial attention only: $SA_{30}$    & {78.54}  & {69.55} & 6.1  \\ 
\hline
Channel attention only: $CA_{12}$   & {91.44}  & {74.88} & 6.4 \\ 
Channel attention only: $CA_{20}$   & {91.53}  & {76.12} & 6.4 \\ 
Channel attention only: $CA_{30}$   & {94.00}  & {77.54} & 6.4 \\ 
\hline
  $sigmoid$ spatial attention: $SA_{12}$  & 77.62  & 70.31 & 3.9\\ 
 $sigmoid$ spatial attention: $SA_{20}$  & {93.06}  & {84.89} & 8.9 \\ 
  $sigmoid$ spatial attention: $SA_{30}$  & 93.90  & {85.96} & 10.2\\ 
\hline
    $P_{12}$ with general dropout  & {92.46} & {84.24}  & 7.9\\ 
   $P_{20}$ with general dropout  & {93.85} & {84.83} & 8.9 \\ 
    $P_{30}$ with general dropout  & 94.24  & {86.61}  & 10.2 \\ 
   \hline 
   $P_{12}$ without $Gaussian$ dropout  & {93.06}  & {83.29} & 7.9 \\ 
  $P_{20}$ without $Gaussian$ dropout  & {94.13}  & {84.60} & 8.9 \\ 
   $P_{30}$ without $Gaussian$ dropout  & {94.82}  & {85.60} & 10.2 \\ 
\hline
 $P_{30}$ with 1 erased region, rand RGB  &94.48 & 84.89 & 10.2 \\
 $P_{30}$ with 1 erased region, RGB=127  &93.69 & 84.13 & 10.2 \\
 $P_{30}$ with 2 erased regions, RGB=127  &94.52 & 86.43 & 10.2 \\
\hline
  $P_{30}$, 2 erased regions, rand RGB: \textbf{SYD-Net}   & \textbf{94.78}  & \textbf{87.73}  & 10.2 \\   
\hline
\end{tabular}
\end{center}
\vspace{-0.6cm}
\end{table}

\subsubsection{Different activations in  MLP and dropout layers}
The effectiveness of $softmax$ over $sigmoid$ activation in the MLP layer of spatial attention  is investigated. It is noted that \textit{softmax} is more efficient in activating the neurons to estimate the probability maps for producing spatial attention masks. However, both activations are useful for improving overall accuracies  leveraging the attention mechanism.
The contribution of the \textit{Gaussian} dropout (GD) is tested over the general dropout for regularization. It is evident that GD improves the learning task and  enhances accuracy.

\subsubsection{Variations in random erasing data augmentation}
The performance of two randomly selected regions over  a single region on input image during image augmentation is tested. The examples of random region erasing are illustrated in Fig. \ref{fig:patches}. The regions are non-overlapping, and the randomness of related hyper-parameters (\textit{e.g.}, size and color) of both regions are independent.  SYD-Net is trained from scratch in two different erasing cases, \textit{i.e.}, one with the random RGB values and another with a fixed RGB=127 value. In this test, we considered only $P_{30}$ patches using MobileNet-v2. It implies that two erased regions could improve the accuracy compared to one erased region in both  cases of RGB values. Because two smaller erased regions can learn more effectively than a larger erased region within the input image and improve the recognition accuracy. Also, random RGB performs slightly better than a fixed value RGB=127. The data augmentation of two erased regions with random RGB values performs more effectively in SYD-Net. Finally, the best model components of SYD-Net rendering the highest performance underlying MobileNet-v2 are given for  completeness of ablation studies, implying the suitability of major components of the proposed SYD-Net architecture. 

\section{Conclusion}  \label{con}
This paper proposes a new patch-based attention method, called SYD-Net for fine-grained human posture recognition. We have introduced and benchmarked two new image datasets, representing  12-dance, and 102-sport actions with diversity. SYD-Net has achieved better performances on the Yoga-82 and Yoga-107 datasets. SYD-Net integrates fixed-size and multi-scale patches to learn contextual information and semantic understanding to define a comprehensive feature descriptor through  spatial and channel attention. Random region erasing data augmentation also improves  accuracy. Overall evaluation of various key components justifies the contribution of each module of SYD-Net. In the future, we plan to develop larger datasets on Sport and Dance styles and explore graph-based deep architecture for human posture recognition.

\section*{Acknowledgement}
The work and the contribution were also supported by the SPEV project, University of Hradec Kralove, Faculty of Informatics and Management, Czech Republic (ID: 2102–2023), “Smart Solutions in Ubiquitous Computing Environments”. We are also grateful for the support of student Michal Dobrovolny in consultations regarding application aspects. We  thank to the Associate Editor and Reviewers for their comments to improve this paper. We thank to M. Verma \cite{verma2020yoga}, and repositories for providing their datasets used in this work. We are thankful to the BITS Pilani,  Pilani Campus, Rajasthan, India, for providing the necessary infrastructure and Research Initiation Grant to carry out this work. 
\ifCLASSOPTIONcaptionsoff
  \newpage
\fi

\bibliographystyle{IEEEtran}
\bibliography{Ref.bib}

\end{document}